\documentclass[manuscript,screen]{acmart}

\usepackage{amsmath,amsfonts}
\usepackage{algorithmic}
\usepackage{algorithm}
\usepackage{array}
\usepackage{textcomp}
\usepackage{stfloats}
\usepackage{url}
\usepackage{verbatim}
\usepackage{graphicx}
\usepackage{booktabs}
\usepackage{threeparttable}
\usepackage{makecell}
\usepackage{color, multirow, amsmath, amsthm, mathrsfs}
\usepackage{bbding}
\usepackage{fontawesome}
\usepackage{makecell}
\usepackage{colortbl}
\usepackage{subfigure}
\usepackage{makecell}
\usepackage{tabularx}
\AtBeginDocument{%
  \providecommand\BibTeX{{%
    \normalfont B\kern-0.5em{\scshape i\kern-0.25em b}\kern-0.8em\TeX}}}

\setcopyright{acmcopyright}
\copyrightyear{2023}
\acmYear{2023}
\acmDOI{XXXXXXX.XXXXXXX}

\acmJournal{JACM}
\acmVolume{37}
\acmNumber{4}
\acmArticle{111}
\acmMonth{8}

\newcommand{\ie}{\textit{i}.\textit{e}.}
\newcommand{\et}{\textit{e}\textit{t} \textit{a}\textit{l}.}
\newcommand{\eg}{\textit{e}.\textit{g}.}

\begin{document}

\title{Computational Analysis of Degradation Modeling in Blind Panoramic Image Quality Assessment}


\authorsaddresses{
Authors' addresses: J. Yan, Z. Tan, J. Rao, L. Wu, Y. Zuo, and Y. Fang (Corresponding author), Jiangxi University of Finance and Economics, Nanchang, China; emails: yanjiebin@jxufe.edu.cn, ziwentan@foxmail.com, jialerao@foxmail.com, leiwu001@foxmail.com, yifanzuo@foxmail.com, fa0001ng@e.ntu.edu.sg.}

\author{Jiebin Yan}
\email{jiebinyan@foxmail.com}
\affiliation{%
 \institution{Jiangxi University of Finance and Economics}
 \city{}
 \country{China}
}

\author{Ziwen Tan}
\email{}
\affiliation{%
  \institution{Jiangxi University of Finance and Economics}
  \city{Nanchang}
  \country{China}
}

\author{Jiale Rao}
\email{}
\affiliation{%
  \institution{Jiangxi University of Finance and Economics}
  \city{Nanchang}
  \country{China}
}

\author{Lei Wu}
\email{}
\affiliation{%
  \institution{Jiangxi University of Finance and Economics}
  \city{Nanchang}
  \country{China}
}

\author{Yifan Zuo}
\email{}     
\affiliation{%
  \institution{Jiangxi University of Finance and Economics}
  \city{Nanchang}
  \country{China}
}

\author{Yuming Fang}
\email{}
\affiliation{%
  \institution{Jiangxi University of Finance and Economics}
  \city{Nanchang}
  \country{China}
}

\renewcommand{\shortauthors}{J. Yan et al.}

\begin{abstract}
Blind panoramic image quality assessment (BPIQA) has recently brought new challenge to the visual quality community, due to the complex interaction between immersive content and human behavior. Although many efforts have been made to advance BPIQA from both conducting psychophysical experiments and designing performance-driven objective algorithms, \textit{limited content} and \textit{few samples} in those closed sets inevitably would result in shaky conclusions, thereby hindering the development of BPIQA, we refer to it as the \textit{easy-database} issue. In this paper, we present a sufficient computational analysis of degradation modeling in BPIQA to thoroughly explore the \textit{easy-database issue}, where we carefully design three types of experiments via investigating the gap between BPIQA and blind image quality assessment (BIQA), the necessity of specific design in BPIQA models, and the generalization ability of BPIQA models. From extensive experiments, we find that easy databases narrow the gap between the performance of BPIQA and BIQA models, which is unconducive to the development of BPIQA. And the easy databases make the BPIQA models be closed to saturation, therefore the effectiveness of the associated specific designs can not be well verified. Besides, the BPIQA models trained on our recently proposed databases with complicated degradation show better generalization ability. Thus, we believe that much more efforts are highly desired to put into BPIQA from both subjective viewpoint and objective viewpoint.      

\end{abstract}





\maketitle


\section{Introduction}\label{sec:intro}
Panoramic image (also called omnidirectional image or 360$^\circ$ image) is able to capture 360 range visual content and provides an immersive experience, which differs substantially from 2D planar image. Accordingly, the interaction between omnidirectional content and users makes the modeling of actual users' viewing behavior in an immersive environment much more complex than planar images, as well as for other relevant tasks, \eg, panoramic image quality assessment (PIQA). Image quality assessment (IQA) has been a basic and classic research topic in the image processing community, which has thrived in a diverse direction in recent years. R\&R-Net~\cite{ma2021remember} focuses on addressing catastrophic forgetting in cross-task blind IQA by introducing a scalable incremental learning framework~\cite{ma2023forgetting}. LIQE~\cite{zhang2023blind} enhances the evaluation ability with the help of multitask learning strategy. DGQA~\cite{li2024bridging} is dedicated to bridging the gap between synthetic and authentic images. IQA also gained much attention from both academia and industry due to its wide applications in algorithm improvement~\cite{fang2021superpixel} and system optimization~\cite{siekkinen2018can,le2021perceptually}.  Compared to other long-standing research topics, \ie, quality assessment of various contents, including general 2D planar image, underwater image~\cite{hou2023uid2021}, computer graphics image~\cite{zhang2023subjective}, stereoscopic image~\cite{fang2019stereoscopic,yan2020blind}, synthesized image~\cite{yan2020no}, algorithm-generated image~\cite{fang2020blind}, 3D point cloud~\cite{zhou2022blind}, \emph{etc.}, PIQA shows particular features in quality degradation modeling, \ie, the internal geometric characteristic of panoramic images and external diverse human viewing behaviors. 

\begin{figure}[]
\centering
\includegraphics[width=0.6\linewidth]{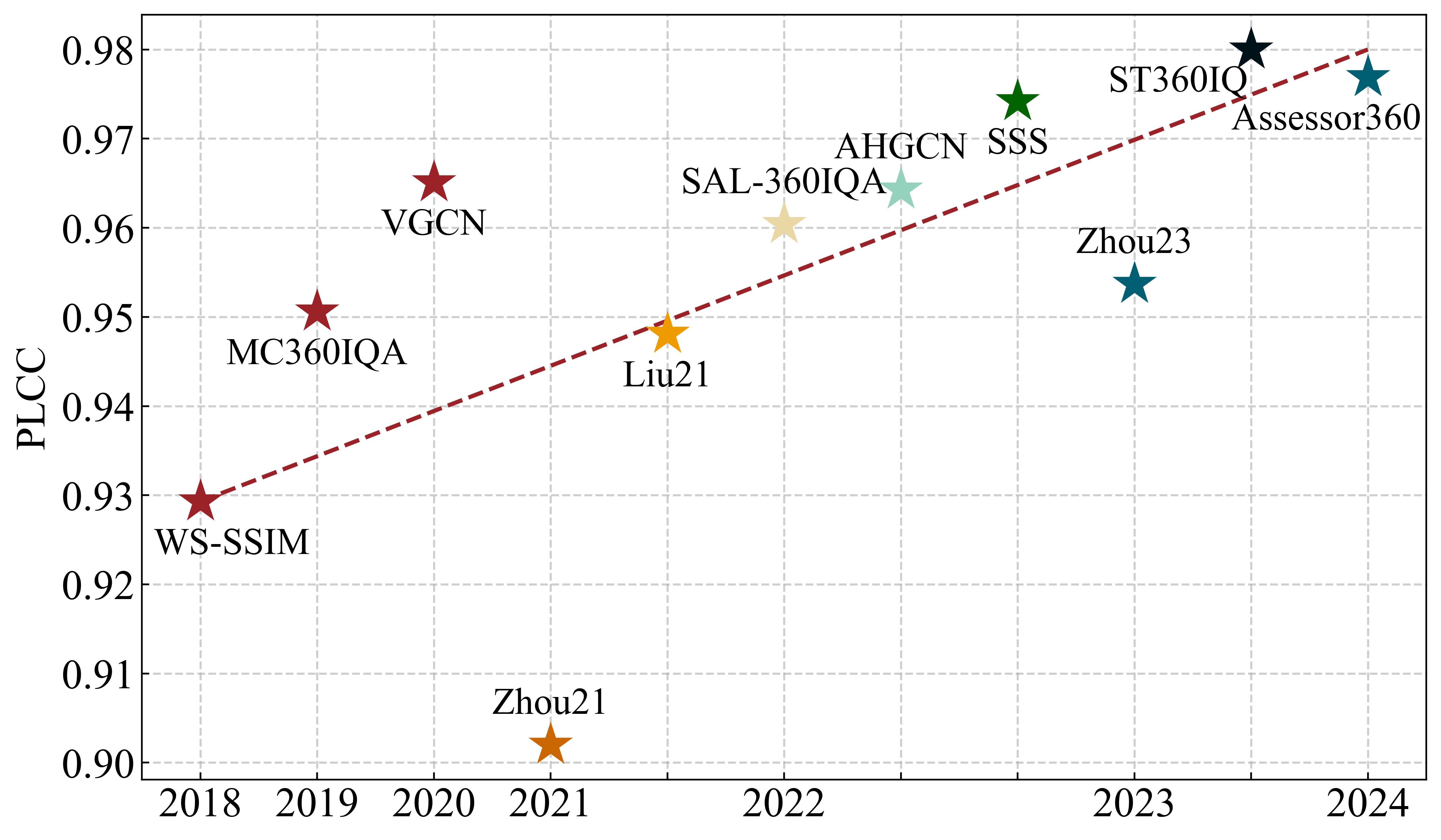}
\caption{An illustration of the advance of PIQA in terms of pearson linear correlation coefficient on the CVIQ database~\cite{sun2018large}. The listed PIQA models include WS-SSIM~\cite{zhou2018weighted}, MC360IQA~\cite{sun2019mc360iqa}, VGCN \cite{xu2020blind}, Zhou21~\cite{zhou2021omnidirectional}, Liu21 \cite{liu2021blind}, SAL-360IQA~\cite{sendjasni2022sal}, AHGCN~\cite{fu2022adaptive}, SSS~\cite{liu2022hvs}, Zhou23~\cite{zhou2023perception}, ST360IQ~ \cite{tofighi2023st360iq}, and Assessor360~\cite{wu2023assessor360}. Note that the phenomenon of performance saturation on the CVIQ database also suits other PIQA databases.}
\label{fig:cgp}
\end{figure}

A straightforward methodology to design PIQA models is to modify widely used full-reference (FR) IQA (FR-IQA) models such as peak signal-to-noise ratio (PSNR) and SSIM~\cite{wang2004image} by considering the aforementioned internal and external features, \eg, 
WS-PSNR~\cite{sun2017weighted}
and WS-SSIM~\cite{zhou2018weighted}, however, they show sub-optimal performance due to their limited quality representation ability and the complexity of modeling human viewing behaviors. Another approach is converting panoramic images into videos by extracting consecutive viewports~\cite{sui2021perceptual}, achieving quality prediction through video quality assessment (VQA) models while using data compression strategy~\cite{yan2024video} to improve process efficiency. In the past few years, PIQA has gained significant development by integrating the characteristics of the human visual system (HVS) and advanced architectures in designing quality degradation modeling modules, especially those deep learning based blind PIQA (BPIQA) models (which can be deployed without access to reference images, and thus more practical in real applications). As shown in Fig.~\ref{fig:cgp}, the recently proposed PIQA models, \ie, ST360IQ~\cite{tofighi2023st360iq} and Assessor360~\cite{wu2023assessor360}, largely surpass these models proposed several years ago, and approximate to 0.98 in terms of pearson linear correlation coefficient on the CVIQ database~\cite{sun2018large}. Note that this situation also exists on other public PIQA databases. From those public results, it seems that the performance of current state-of-the-art PIQA models has been saturated, and thus only leaves limited room for improvement. Actually, as this regard, the core issue, \ie, data, has been overlooked. As shown in Tab.~\ref{tab:databases_summary}, one of the largest so far public databases contains only slightly more than one thousand panoramic images, which is very sparse in the whole panoramic image space. Compared to those public IQA databases~\cite{ma2016waterloo,fang2020perceptual}, these public PIQA databases (except the JUFE-10K and OIQ-10K databases~\cite{yan2024subjective,yan2024omnidirectional}) also suffer from single image content and simple degradation situation. This embarrassing fact is in stark contrast to the data-hungry characteristic of deep learning models, \ie, the limited number of data brings high-risk over-fitting problem and therefore reduces the credibility of the comparison results on those public PIQA databases. Besides, the repetitive training on these closed sets would easily lead to data leakage. By considering this, we argue that the \textit{real} development status should be overshadowed by the shortage of large-scale public PIQA databases.

\begin{table*}
\setlength{\tabcolsep}{4pt}
\small
\centering
\caption{The Summary of Existing PIQA Databases. Note that these Two Databases Marked by ``$\lozenge$" are our Newly Constructed Databases for Exploring the Heterogeneous Degradation Problem in PIQA. ``Homo" and ``Hetero" Represent Homogeneous and Heterogeneous Distortion, Respectively. ``GB" Represents Gaussian Blur. ``GN" Represents Gaussian Noise. ``WN" Represents White Noise. "BD" Represents Brightness Discontinuity. ``ST" Represents Stitching. "Projection" Represents Distortion with the Mapping Method. ``TM" Represents Tone Mapping. ``DS" Represents Downsampling. ``AU" Represents Authentic Distortion.}
\begin{tabular}{lcccccccc}
\toprule
\textbf{Database} & \textbf{\# Year} & \textbf{\# Images} & \textbf{Resolution} & \textbf{Degradation} & \textbf{Distortion Type} & \textbf{Distortion Level} \\
\midrule
CVIQ \cite{sun2018large} & 2018 & 544 & 4K & Homo & JPEG, AVC, HEVC & 11 \\
OIQA \cite{duan2018perceptual} & 2018 & 336 & 11K, 13K & Homo & JPEG, JP2K, GB, GN & 5 \\
MVAQD \cite{jiang2021cubemap} & 2021 & 315 & 4K$\sim$12K & Homo & JPEG, JP2K, HEVC, WN, GB & 4 \\
IQA-ODI \cite{yang2021spatial} & 2021 & 1,080 & 8K & Homo & JPEG, Projection & 4 \\
JUFE \cite{fang2022perceptual} & 2022 & 1,290 & 8K & Hetero & GB, GN, BD, ST & 3 \\
OSIQA~\cite{duan2023attentive} & 2023 & 300 & 6K & Hetero & ST, Projection & 4 \\
AIGCOIQA \cite{yang2024aigcoiqa2024} & 2024 & 300 & 4K & AIGC & - & - \\
JUFE-10K \cite{yan2024subjective}~$\lozenge$ & 2024 & 10,320 & 8K & Hetero & GB, GN, BD, ST & 3 \\
OIQ-10K \cite{yan2024omnidirectional}~$\lozenge$ & 2024 & 10,000 & 0.8K$\sim$20K & Homo, Hetero & \makecell[c]{JPEG, JP2K, JPEG XT, TM, AVC, \\HEVC, VP9, GB, GN, BD, ST, DS, AU} & - \\
\bottomrule
\end{tabular}
\label{tab:databases_summary}
\end{table*}

Motivated by these studies regarding model evaluation~\cite{yan2021exposing,fang2023study,sun2024analysis}, in this paper, we aim to seek the \textit{truth} of current status of BPIQA, and discuss the strengths and weaknesses of these BPIQA models to find potential directions for next-generation subjective databases and objective models. We start from the data issue in PIQA (a.k.a, the \textit{easy-database} problem~\cite{sun2024analysis}), it highlights that closed sets with limited content and few samples cannot fully realize the potential of BPIQA models, thereby hindering the development of BPIQA. Therefore, three perspectives to be investigated naturally arise: the gap between BPIQA and blind image quality assessment (BIQA), the necessity of specific design in BPIQA model, and the generalization ability of BPIQA models. For the first perspective, following the current mainstream design concept, \ie, extract viewports as models' input like users' viewing process, we use the (weighted) outputs of an IQA model on viewports as the quality prediction of the corresponding panoramic image. For the second perspective, we try to design a family of BPIQA models with only \emph{pure} backbone network (\ie, without any specific designs like PIQA models), and compare them with current deep learning based BPIQA, BIQA and VQA models. For the third perspective, we examine it by cross-database validation, verifying whether the test BPIQA models overfit in the public PIQA databases and therefore show unreal prosperity. 

In summary, our contributions are three folds:
\begin{itemize}

\item We present the first computational analysis to uncover the truth of current development status of BPIQA, and point out the design direction of next-generation BPIQA models. 

\item We take a close look at BPIQA (deep learning models) from three elaborate perspectives, including the gap between BPIQA and BIQA, the necessity of specific design in BPIQA models, and the generalization ability of BPIQA models. 

\item We conduct detailed quantitative experiments and give qualitative analysis, where we find that some public PIQA databases seem to be relatively simple for these deep learning based BPIQA models, and much more efforts are desired to be put in the future. 

\end{itemize}

\section{Related Work}
\label{sec:rw}

In this section, we first introduce objective BPIQA models, and then describe the studies regarding model evaluation.

\subsection{Objective BPIQA Models}
\label{subsec:objective}

BPIQA is a non-trivial task since it can be deployed in real time dynamic bandwidth allocation~\cite{liu2023quality}. Same to BIQA models, those hand-crafted features based BPIQA models~\cite{liu2023toward,jiang2021cubemap} follow a general two-independent-step framework, including quality-sensitivity feature extraction by considering prior knowledge and quality regression by a shallow learning algorithm. Limited by the representation ability and learning capacity, those models have been gradually surpassed by deep learning based BPIQA models, which can be categorized into viewport-unaware BPIQA models~\cite{xu2020blind,sun2019mc360iqa,zhou2021omnidirectional,chai2021monocular,zhou2023perception} and viewport-aware BPIQA models~\cite{fang2022perceptual,yang2022tvformer,wu2023assessor360,liu2024perceptual} (we omit \textit{deep learning based} for clarity). Note that viewport-unaware means these models measure quality degradation of panoramic images without considering human browsing behavior (\ie, only a limited field of view can be seen by users at any single moment), and vice versa. Some of these viewport-unaware BPIQA models~\cite{xu2020blind,chai2021monocular} accept the \textit{originally distorted} panoramic image, \eg, in the format of equirectangular projection (ERP), as input. Xu~\et~\cite{xu2020blind} built a spatial viewport graph to capture the dependency between local viewports and a global branch to predict the global quality of the entire panoramic image. Chai~\et~\cite{chai2021monocular} designed a three-channel network to simulate the complex interaction of monocular and binocular perception, and introduced deformable convolution to process non-uniform sampling distortion distributions. The others~\cite{sun2019mc360iqa,zhou2021omnidirectional,zhou2023perception} extract viewports from different directions, while also ignoring human behavior when browsing panoramic images.

As shown in our previous study~\cite{sui2021perceptual}, viewing conditions, including the starting point and the exploration time, have significant influence on users' quality of experience. Considering this, Fang~\et~\cite{fang2022perceptual} took an initial step to an end-to-end BPIQA model by injecting the ground-truth starting point (which is represented by its coordinates) and the exploration time to quality-aware feature representation. However, this model~\cite{fang2022perceptual} is unfriendly when user's viewing behavior is unknown. Liu~\et~\cite{liu2024perceptual} recently extended this BPIQA model~\cite{fang2022perceptual} by substituting the backbone with a more powerful encoder. Yang~\et~\cite{yang2022tvformer} designed a trajectory-guided BPIQA model, which is jointly optimized by head trajectory prediction and quality assessment. Wu~\et~\cite{wu2023assessor360} proposed a recursive probability sampling method with the consideration of content and detailed information for generating viewport sequences, and input these sequences to the multi-scale feature aggregation module for capturing viewport-wise distortion and followed by a temporal modeling module for learning the viewport transition. Recently, Yan~\et~\cite{yan2024multitask} introduced a multi-task auxiliary network framework designed to assess heterogeneously distorted panoramic images by emulating the parallel learning capabilities of the human brain.

To sum up, current BPIQA models achieve significant performance on the homogeneous PIQA databases (\textit{e.g.}, CVIQ and OIQA). However, most of them have not been tested on large-scale PIQA databases and overlooked heterogeneous distortions, which obscures the \textit{real} development status of PIQA. In this work, we conduct extensive quantitative experiments, focusing on analyzing model performance across different databases to reveal the current state of PIQA development.

\subsection{Model Evaluation}
\label{subsec:model}

Instead of technically pursuing \textit{high-performance} on those closed sets, these works regarding model evaluation aim to look back over what we have done in certain research topics and reflect on what should we do next. To this end, a natural and common way is to revisit the failure sources which make models get into trouble~\cite{ma2020group,wang2020i,yan2021exposing}. Zhang~\et~\cite{zhang2016far} compared a top-performing single frame pedestrian detector with a human baseline, and identified both localisation and background-versus-foreground errors manually. Besides, the authors addressed these two types of errors by studying the impact of training annotation noise and convents for pedestrian detection, respectively.
Kamann~\et~\cite{kamann2020benchmarking} exploited an extensive evaluation of the robustness of semantic segmentation against real-world corruptions. Other than 
corruptions~\cite{kamann2020benchmarking} or exposing failures~\cite{ma2020group,wang2020i,yan2021exposing}, Recht~\et~\cite{recht2019imagenet} found that the performance of ImageNet~\cite{deng2009imagenet} classifiers drops range from 3\% to 15\% on two prominent benchmarks,~\ie, CIFAR-10 and ImageNet. Speculated from the results, the drops are resulted from the models' helplessness to generalize the slightly ``harder'' samples than that in original test sets.

The aforementioned studies are conducted in the scope of closed sets, which generally refer to the held-out test sets of the public databases. In~\cite{wang2008maximum}, Wang~\et~proposed a novel model comparison method namely MAximum Discrepancy (MAD) competition from the perspective of ``analysis by synthesis'', which only needs few samples. In the concept of MAD, the model who has stronger ability to falsify the other model is considered better. The pioneer MAD competition methodology, however, requires the compared models to be differentiable, and thus is hard to extend,~\eg, comparing the non-differentiable models. Ma~\et~\cite{ma2020group} extended this work~\cite{wang2008maximum} to an easily extensible methodology for comparing multiple computational models regardless of whether the compared models are differentiable or not. The concept of MAD competition has also been successfully adapted to IQA~\cite{wang2021active,wang2021troubleshooting}, image classification~\cite{wang2020i}, and semantic segmentation~\cite{yan2021exposing}. 

Although numerous research studies have been conducted on model evaluation, such as IQA, image segmentation, and object detection, there has yet to be a similar systematic review in the PIQA domain. To address the potential shortcomings in existing PIQA research, we conduct a detailed study on PIQA by considering different model evaluation approaches.

\section{The Generic BPIQA Framework}
\label{sec:framework}
In this section, we describe the generic BPIQA framework, which includes a viewport generator for simulating human behavior, a viewport-wise feature extractor for capturing intra-viewport quality degradation, an inter-viewport interaction module for measuring their collective influence, and a quality regression module for mapping quality-sensitive features to global quality.

\subsection{Viewport Generator}
\label{subsec:viewport_generator}
Given $N$ panoramic images $\{PI_n\}_{n=1}^N$, where $PI_n\in \mathbb{R}^{H\times W\times 3}$ denotes the $n$-th panoramic image, and $H$ and $W$ denote its height and width, respectively. Firstly, $M$ viewports of $n$-th panoramic image are selected by a sampling strategy:
\begin{align}
\label{eq:qcr}
  \mathcal{V}=f_s(PI_n;\xi),
\end{align}
where $\mathcal{V}=\{\textbf{v}_n^m\}_{m=1}^M$;$\textbf{v}_n^m\in \mathbb{R}^{H_v\times W_v\times 3}$ denotes the $m$-th extracted viewport of the $n$-th panoramic image, and $H_v<H$, $W_v<W$; $f_s$ and $\xi$ denote the sampling operation and corresponding sampling trajectory, respectively. Instead of using default scanpath~\cite{sui2021perceptual,zhang2022no} or predicting scanpath~\cite{sui2023perceptual}, we adopt a simple yet efficient way by considering that users are more likely to view the regions along the equator of panoramic image~\cite{deng2021lau}, which can be formulated as follows:   
\begin{align}
\label{eq:vp_sampling}
  \textbf{v}^m_n=PI_n[(\phi+A^m_n, 0);FoV],
\end{align}
where $[\cdot;\cdot]$ is the sampling operation according to coordinate and field of view (FoV), FoV=$\pi$/3; $\phi$ denotes the start point whose default value is 0, and $A_n^m$ denotes the longitude offset of the $m$-th viewport of the $n$-th panoramic image. In this paper, $A_n^m$ is set to 45° and eight viewports with size of 224$\times$224 are extracted.

\subsection{Viewport-wise Feature Extractor}
\label{subsec:vp_feat_ex}
Similar to BIQA and VQA models, the feature extractor $f_e(\cdot)$ in BPIQA models takes a viewport $\textbf{v}$ as input, and extracts corresponding quality-aware features $\textbf{x}$, which can be formulated as follows: 
\begin{align}
\label{eq:qcr}
  \textbf{x}^m_n=f_e(\textbf{v}^m_n;\theta_e), \text{for}~m=1,2,\cdots, M,
\end{align}
where $\textbf{x}^m_n\in \mathbb{R}^{H_x\times W_x\times C}$ denotes the extracted features, and $H_x<H_v$, $W_x<W_v$; $\theta_e$ denotes the learnable parameters. Note that $f_e$ can be instantiated by any convolutional neural network (CNN), Transformer or their hybrid architecture. In practice, we remove the head layer and sequentially process the viewports through the network $f_e$ to obtain visual perceptual features. The impact of different types of feature extractors is discussed in Section~\ref{subsec:analysis}.

\subsection{Inter-viewport Interaction Module}

This module aims to measure the collective influence of selected viewports on panoramic image quality, which plays a significant importance in mimicking users' viewing behavior~\cite{hands2001recency,fang2022perceptual}, and can be formulated as follows:
\begin{align}
\label{eq:qcr}
  \bar{\textbf{x}}_n=f_a(\textbf{x}_n;\theta_a),
\end{align}
where $\bar{\textbf{x}}_n\in \mathbb{R}^{D\times M}$ denotes the merged quality features of $n$-th panoramic image, $D$ and $M$ denote the dimension of $\bar{\textbf{x}}_n$ and the number of viewports, respectively; $\textbf{x}_n=[\textbf{x}^1_n,\textbf{x}^2_n,\cdots,\textbf{x}^M_n]$; $f_a$ and $\theta_a$ denote the interaction module and its parameters, respectively. Recurrent neural networks, similar to those used in VQA, are commonly employed to capture interaction dependencies between viewports during the viewing process, it can be formulated as follows:
\begin{align}
\label{eq:rnn}
  h_t=\sigma (\textbf{W}_hh_{t-1}+\textbf{W}_x\textbf{x}_n^m+b),
\end{align}
where $h_{t-1}$ and $h_t$ represent the previous hidden state and the current hidden state, respectively; $\sigma$ is the Sigmoid activation function, and $b$ is the bias. Note that the detailed analysis regarding its effectiveness will be given in Section~\ref{subsec:analysis}.

\subsection{Quality Regression Module}
\label{subsec:quality_res}
Finally, the quality score $s_n$ of $n$-th panoramic image can be obtained by a quality regression operation:
\begin{align}
\label{eq:qcr}
  s_n=f_r(\bar{\textbf{x}}_n;\theta_r),
\end{align}
where $f_r$ and $\theta_r$ denote the regression method and its parameters, respectively. In previous works, there are mainly two regression methods: integrated mapping and partial mapping, which will be discussed in Section~\ref{subsec:analysis}. The former utilizes interaction modeling to fuse the features of different viewports from the same panoramic image, and then the quality score $s$ can be mapped by several fully connected (FC) layers. While in the latter one, the features $\textbf{x}_n^m$ of the $m$-th viewport in the $n$-th panoramic image are first mapped to a quality score:
\begin{align}
\label{eq:qcr}
  s_n^m=\textbf{P}\textbf{x}_n^m+b, \text{for}~m=1,2,\cdots,M,
\end{align}
where $\textbf{P}\in \mathbb{R}^{1\times D}$ and $b$ denote the learnable projection layer and bias, respectively. Subsequently, these quality scores are then aggregated using average pooling:
\begin{align}
\label{eq:qcr}
  s_n=\frac{1}{M}\sum_{m=1}^M s_n^m.
\end{align}

\section{Experiments}

In this section, we first introduce the test PIQA databases. Then, we describe the experimental settings in detail, including the proposed BPIQA models and the evaluation criteria. Later, a quantitative analysis is conducted to explore the \textit{easy-database} issue in BPIQA. Finally, we summarize the experimental results.

\begin{figure*}[]
\centering
\subfigure[]{
\includegraphics[width=0.25\linewidth]{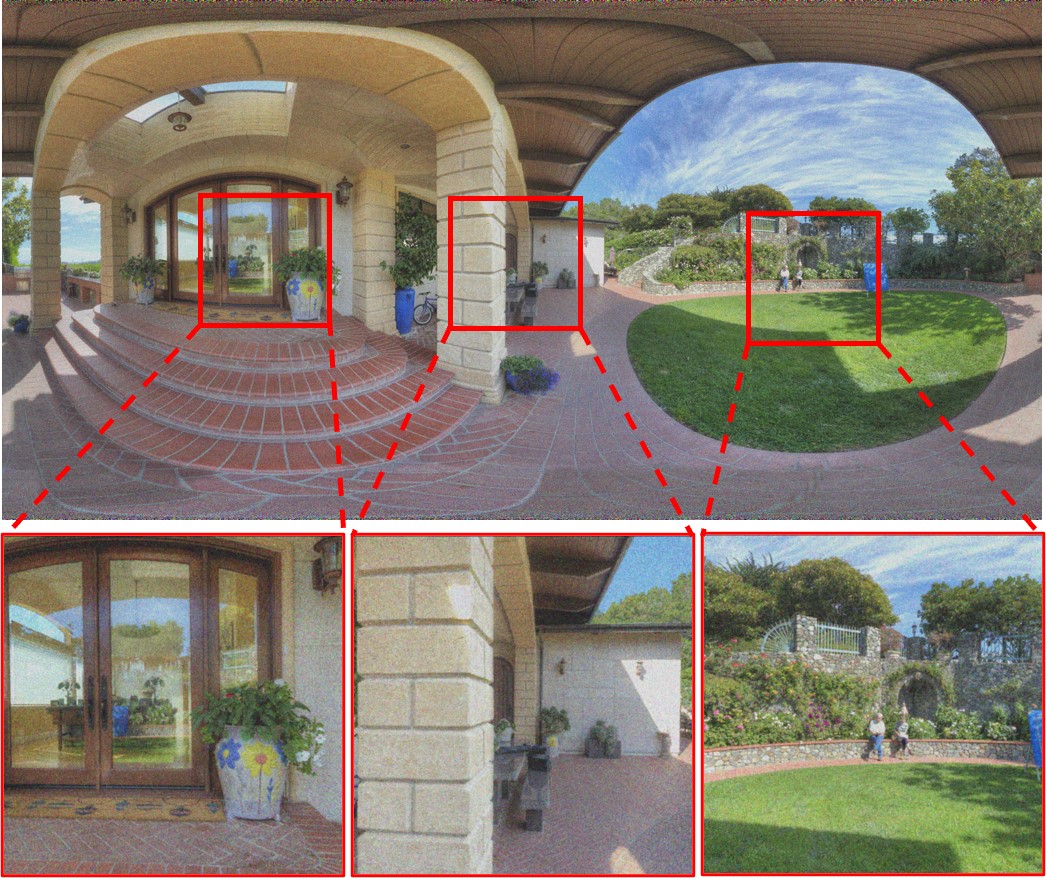}
}
\subfigure[]{
\includegraphics[width=0.25\linewidth]{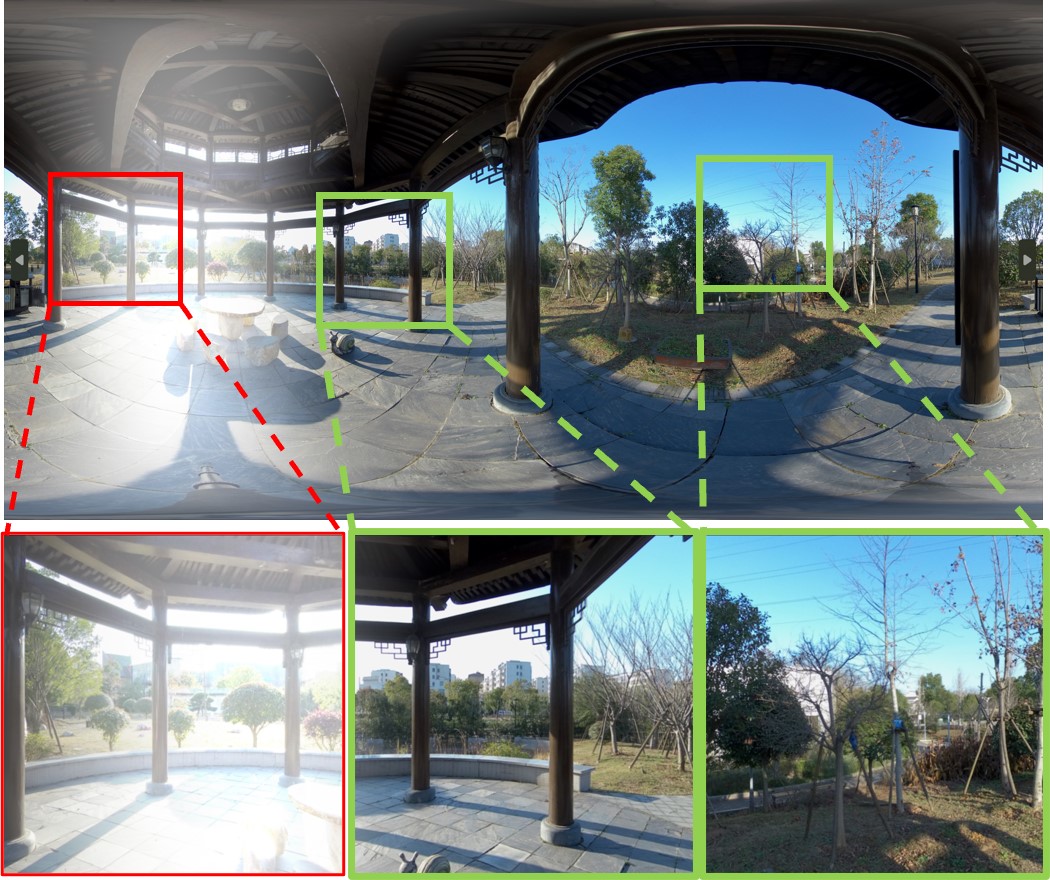}
}
\subfigure[]{
\includegraphics[width=0.25\linewidth]{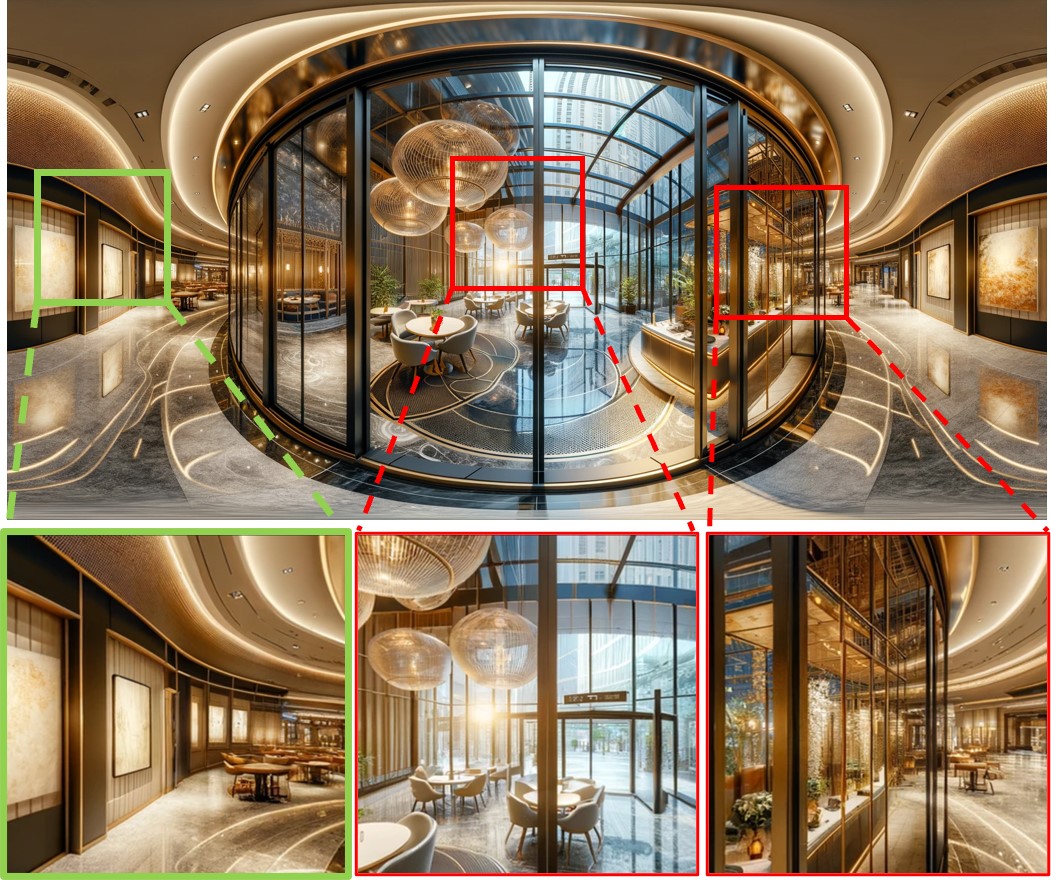}
}
\caption{The visual examples of (a) homogeneously distorted panoramic image from OIQA database \cite{duan2018perceptual}, (b) heterogeneously distorted panoramic image from OIQ-10K database \cite{yan2024omnidirectional} and (c) generated panoramic image from AIGCOIQA database \cite{yang2024aigcoiqa2024}. Note that viewports outlined in green indicate high visual quality, while those with red contours signify low visual quality.}
\label{fig:dis_type}
\end{figure*}

\subsection{The Test PIQA Databases.} 
We conduct experiments on eight public PIQA databases, including CVIQ~\cite{sun2018large}, OIQA~\cite{duan2018perceptual}, MVAQD~\cite{jiang2021cubemap}, IQA-ODI~\cite{yang2021spatial}, OSIQA~\cite{duan2023attentive}, OIQ-10K~\cite{yan2024omnidirectional}, JUFE-10K~\cite{yan2024subjective}, and AIGCOIQA~\cite{yang2024aigcoiqa2024}. According to the distortion situation, these databases can be roughly categorized into three classes: \textit{homogeneous distortion}, \textit{heterogeneous distortion}, and \textit{generated distortion}. As shown in Fig.~\ref{fig:dis_type}, the panoramic image with \textit{homogeneous distortion} exhibits similar visual quality in each region, the panoramic image with \textit{heterogeneous distortion} exhibits varied perceptual quality in the viewing process, and the panoramic image with \textit{generated distortion} could show unreasonable elements, \textit{e.g}, suspended lights and wrong light reflection in Fig.~\ref{fig:dis_type} (c). Other details of these databases can refer to Tab.~\ref{tab:databases_summary}.

Note that although the source images in \textbf{CVIQ} contain diverse scenes, the distortion type and content (\textit{e.g.} composition and colorfulness) are relatively monotonous, which may be insufficient to cover the full range of real-world scenarios. Similar to CVIQ, \textbf{OIQA} also contains limited content and a small number of samples, making it impossible to improve the database's quality fundamentally. \textbf{IQA-ODI} focuses on the impairments of JPEG compression and map projection on panoramic images, where the projection patterns include cubemap projection format (CMP), cylindrical projection format (CPP), icosahedron projection format (ISP), and octahedron projection format (OHP).~\textbf{JUFE} is the first large-scale database to study the \textit{heterogeneous distortion} issue, where the disruption is randomly applied to one of the six lenses and then get the locally distorted panoramic images. \textbf{OSIQA} focuses on distortions caused by stitching and includes 300 distorted panoramic images generated from 12 original scenes. To study the unique distortions emerging in AI-generated panoramic images, \textbf{AIGCOIQA} generates 300 panoramic images by leveraging multiple AIGC models and employing 25 textual prompts, and all images have the same resolution of 4K. AIGCOIQA explores the perceptual quality of generated images from three perspectives: quality, comfortability, and correspondence. Each image contains latent \textit{generated distortion} due to the inherent lack of realism in AIGC models. Compared to the JUFE database, \textbf{JUFE-10K} offers more samples, a richer variety of scenes, and more pronounced and diverse quality discontinuities. \textbf{OIQ-10K} is a large-scale database containing 10,000 panoramic images with both \textit{homogeneous distortion} and \textit{heterogeneous distortion}. To deal with the limited diversity in image content and distortion variation, OIQ-10K gathers panoramic images from existing PIQA databases and online websites, covering the vast majority of application scenarios. Four types of distortion range are taken into account, \ie, images without perceptibly distorted region, images with one distorted region, images with two distorted regions, and images with global distortion.

\subsection{Experimental Settings}
As described in Section~\ref{sec:framework}, a basic BPIQA model requires four parts, including a viewport generator, a viewport-wise feature extractor, an inter-viewport interaction module, and a quality regression module. Naturally, BIQA and VQA models can serve as BPIQA models with minor modifications. The former ones could learn quality prior knowledge through pre-training on large-scale planar image quality databases such as KonIQ-10K~\cite{hosu2020koniq}, and the latter ones could cope with users' viewing behavior by utilizing a temporal module. To comprehensively revisit this filed, we test eight BIQA models, including HyperIQA~\cite{su2020blindly}, LinearIQA~\cite{li2020norm}, UNIQUE~\cite{zhang2021uncertainty}, CONTRIQUE~\cite{madhusudana2022image}, MANIQA~\cite{yang2022maniqa}, VCRNet~\cite{pan2022vcrnet}, LIQE~\cite{zhang2023blind}, and QualiCLIP~\cite{agnolucci2024qualityaware}. All BIQA models take eight viewport images as input, and the viewport size is set to $224\times 224$. To obtain the final predicted quality score, each model predicts the quality of each viewport and then averages all scores to a final score. These BIQA models are pre-trained on the KonIQ-10K~\cite{hosu2020koniq} (except that VCRNet is pre-trained on LIVE II~\cite{sheikh2006statistical}, since its pre-trained weights on KonIQ-10K are unavailable). Three VQA models are taken into account to explore the effect of the temporal module in degradation modeling, including VSFA~\cite{li2019quality}, Transformer-VSFA (a variant of VSFA by substituting the backbone with the pure transformer~\cite{vaswani2017attention}), and FAST-VQA~\cite{wu2022fast}. Since VQA models take consecutive frames as input, providing more frames as input is essential. Therefore, we extract 30 viewport images by applying the Equator (\ref{eq:vp_sampling}) where $A$ is set to 12°, to serve as the input for the VQA models. Besides, we test three top-performing BPIQA models, including MC360IQA~\cite{sun2019mc360iqa}, Assessor360~\cite{wu2023assessor360} and OIQAND~\cite{yan2024subjective} to study the impact of the inter-viewport interaction module. To eliminate the design bias in existing BPIQA models, we additionally propose five BPIQA models using only \textit{pure} backbone network as the baseline models, including ResNet-50-P~\cite{he2016deep}, ResNet-152-P~\cite{he2016deep}, Swin-v2-T-P~\cite{liu2022swin}, Swin-v2-B-P~\cite{liu2022swin}, and MaxViT-T-P~\cite{tu2022maxvit} (`-P' denotes the variant for panoramic image). As described in Section~\ref{sec:framework}, we replace the head layer of these general models with the FC layers to obtain the quality score of each viewport, and the average pooling is adopted to fuse local scores to the global score. The summary of these models is shown in Tab.~\ref{tab:models_summary}.

\begin{table*}[]
\centering
\small
\caption{Prevailing Computational Models for BPIQA. ``VPs" Denotes Viewport Images. ``WP" Represents Weighted Pooling.}
\begin{tabular}{lccccccc}
\toprule
\textbf{Model} & \textbf{\# Year} & \textbf{Type} & \textbf{\makecell[c]{Input \\Type}} & \textbf{\makecell[c]{Input \\Stream}} & \textbf{\makecell[c]{Feature \\Extractor}} & \textbf{\makecell[c]{Viewport \\Interaction}} & \textbf{\makecell[c]{Quality \\Regressor}} \\
\midrule
HyperIQA \cite{su2020blindly} & 2020 & BIQA & Image &  8VPs & CNN & None & MLP \\
LinearIQA \cite{li2020norm} & 2020 & BIQA & Image &  8VPs & CNN & None & MLP \\
UNIQUE \cite{zhang2021uncertainty} & 2021 & BIQA & Image &  8VPs & CNN & None & MLP \\
CONTRIQUE \cite{madhusudana2022image}& 2022 & BIQA & Image &  8VPs & CNN & None & MLP \\
MANIQA \cite{yang2022maniqa}& 2022 & BIQA & Image &  8VPs & Transformer & None & WP \\
VCRNet \cite{pan2022vcrnet}& 2022 & BIQA & Image &  8VPs & CNN & None & MLP \\
LIQE \cite{zhang2023blind}& 2023 & BIQA & Image\&Text &  8VPs & Transformer & None & Probability \\
QualiCLIP~\cite{agnolucci2024qualityaware} & 2024 & BIQA & Image\&Text &  8VPs & Transformer & None & Similarity \\
VSFA \cite{li2019quality}& 2019 & VQA & Video &  30VPs & CNN & GRU & WP \\
Transf-VSFA \cite{vaswani2017attention}& 2017 & VQA & Video &  30VPs & Transformer & GRU & WP \\
FAST-VQA \cite{wu2022fast}& 2020 & VQA & Video &  30VPs & Transformer & Video Swin-T & Conv3d\&GAP \\
\midrule
MC360IQA \cite{sun2019mc360iqa}& 2019 & BPIQA & Image &  8VPs & CNN & Integrated Mapping & MLP \\
Assessor360 \cite{wu2023assessor360}& 2024 & BPIQA & Image &  15VPs & Transformer & RSP\&GRU & MLP \\
OIQAND~\cite{yan2024subjective} & 2024 & BPIQA & Image &  8VPs & Transformer & Viewports Vit & MLP \\
ResNet-50-P \cite{he2016deep}& 2016 & BPIQA & Image &  8VPs & CNN & None & MLP \\
ResNet-152-P \cite{he2016deep}& 2016 & BPIQA & Image &  8VPs & CNN & None & MLP \\
Swin-v2-T-P \cite{liu2022swin}& 2022 & BPIQA & Image &  8VPs & Transformer & None & MLP \\
Swin-v2-B-P \cite{liu2022swin} & 2022 & BPIQA & Image &  8VPs & Transformer & None & MLP \\
MaxViT-T-P \cite{tu2022maxvit}& 2022 & BPIQA & Image &  8VPs & Hybrid & None & MLP \\
\bottomrule
\end{tabular}

\label{tab:models_summary}
\end{table*}

In experiments, we randomly split the panoramic images in each PIQA database into 80\% and 20\% for training and testing, respectively. Since each distorted image in the OSIQA database is associated with two MOS values (corresponding to different viewing angles), we use the average of the two MOS values as the overall quality of the image to optimize BPIQA models. The proposed five BPIQA models are pre-trained on the ImageNet~\cite{russakovsky2015imagenet}, and all experiments are implemented by the PyTorch framework~\cite{paszke2017automatic}. In the training phase, these computational models are optimized by the Adam~\cite{kingma2014adam} with an initial learning rate of $10^{-4}$ and the cosine decay learning rate with the minimum learning rate of $10^{-6}$. The loss function is Norm-in-Norm loss~\cite{li2020norm}. The batch size and epochs are set to 16 and 50, respectively. Note that, for models other than those proposed, the training settings are kept consistent with those in the original papers to minimize the influence of other factors. In the cross-database validation, every image from the testing database is used. All experiments are conducted on an NVIDIA GeForce GTX 4090 machine.

\subsection{Performance Evaluation Criteria}
We employ Pearson’s linear correlation coefficient (PLCC) and Spearman’s rank order correlation coefficient (SRCC) as the criteria to evaluate these BIQA and BPIQA models. PLCC measures the accuracy of the prediction, and SRCC measures the consistency in monotonicity between the predicted and true scores, where a higher value denotes better performance. As suggested in~\cite{sheikh2006statistical}, a five-parameter logistic function is applied before calculating PLCC.

\begin{figure*}[]
\centering
\includegraphics[width=0.8\linewidth]{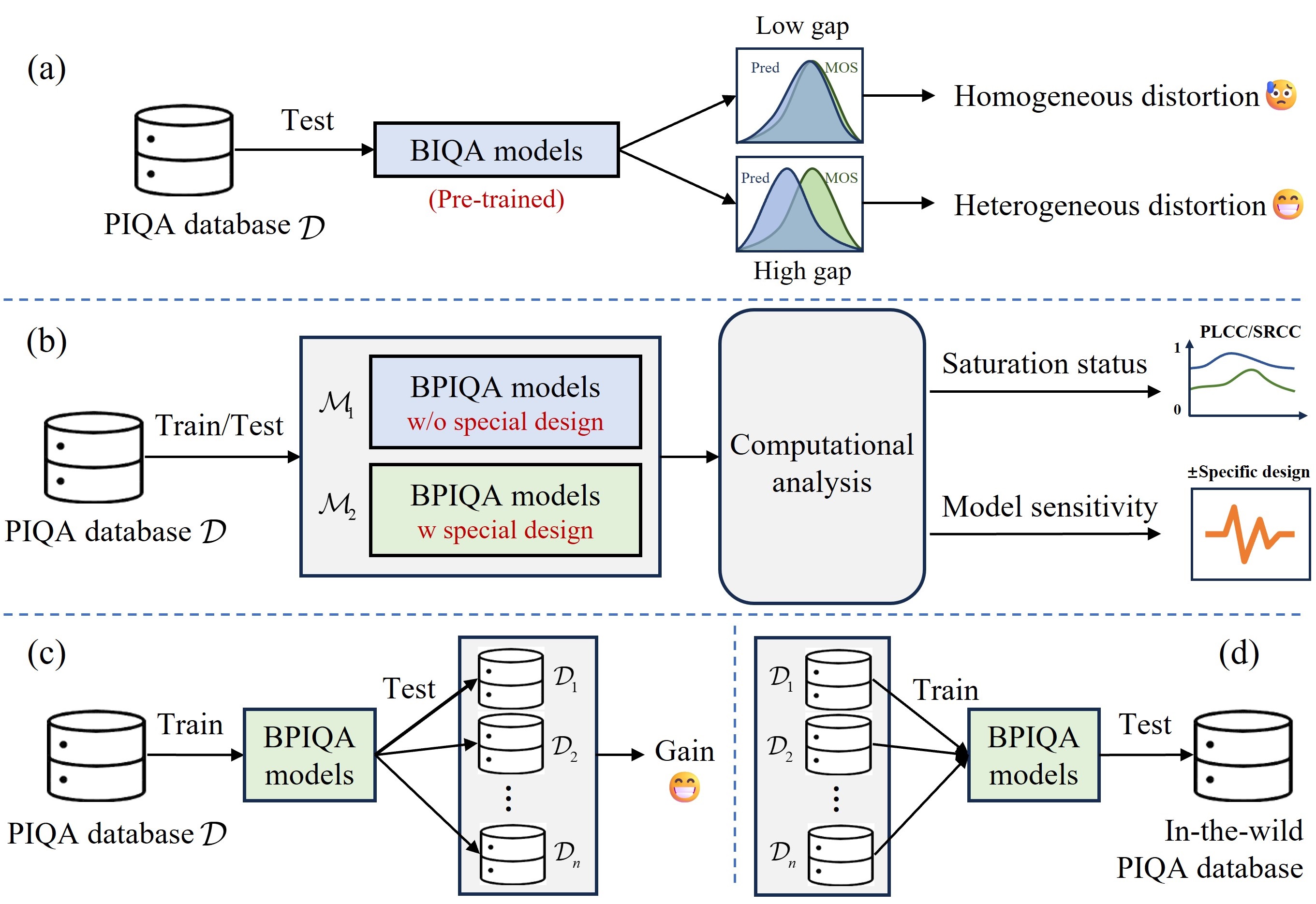}
\caption{The analysis framework for investigating the easy-database issue. We carry out this study from three perspectives: 1) the gap between BPIQA and BIQA (subfig (a)), 2) the necessity of specific design in BPIQA models (subfig (b)), and 3) the generalization ability of BPIQA models (subfigs (c) and (d)).}
\label{fig:analysis_framework}
\end{figure*}

\subsection{Experimental Results and Quantitative Analysis}
\label{subsec:analysis}
To investigate the \textit{easy-database} issue thoroughly, we present an analysis framework, as shown in Fig.~\ref{fig:analysis_framework}. Specifically, we first investigate the gap between BIQA and BPIQA by testing BIQA models on public PIQA databases, which provides a preliminary and rough judgment on whether the test database is conducive to the development of PIQA. To further study whether specific designs contribute to the development of PIQA, we examine the impact of various design elements on evaluation performance. Besides, we conduct experiments across multiple databases to explore how the test database enables BPIQA models to handle challenging samples. This involves a series of cross-database validations and a debiased evaluation. The detailed discussions are as follows.

\begin{table*}[t]
\setlength{\tabcolsep}{1.85pt}
\centering
\small
\caption{Performance Comparison of BIQA Models on These PIQA Databases. The "(PLCC/SRCC)" Denotes the Original Performance in Their Papers. All BIQA Models are Pre-trained on the KonIQ-10K~\cite{hosu2020koniq}, Except that VCRNet is Pre-trained on the LIVE II~\cite{sheikh2006statistical} Database. Note that the Source Database on which QualiCLIP is Pre-trained cannot be Determined, Therefore the Gap cannot be Calculated.}
\begin{tabular}{lcccccccccccccc}
\toprule
  & \multicolumn{2}{c}{CVIQ (7)} & \multicolumn{2}{c}{OIQA (6)} & \multicolumn{2}{c}{MVAQD (5)} & \multicolumn{2}{c}{IQA-ODI (3)} & \multicolumn{2}{c}{OSIQA (1)}& \multicolumn{2}{c}{OIQ-10K (4)} & \multicolumn{2}{c}{JUFE-10K (2)} \\
\multirow{-2}{*}{Method} & PLCC & SRCC & PLCC & SRCC & PLCC & SRCC & PLCC & SRCC & PLCC & SRCC& PLCC & SRCC & PLCC & SRCC  \\
\midrule
HyperIQA & 0.725 & 0.731 & 0.597 & 0.573 & 0.611 & 0.610 & 0.318 & 0.331 & 0.353 & 0.239& 0.589 & 0.563 & 0.215 & 0.208  \\
\rowcolor{gray!30}
Gap (0.917/0.906) &-20.9\%&-19.3\%&-34.9\%&-36.8\%&-33.4\%&-32.7\%&-65.3\%&-63.5\%&\textbf{-61.5\%}&\textbf{-73.6\%}&-35.8\%&-37.9\%&-76.6\%&-77.0\%
 \\
LinearityIQA & 0.747 & 0.744 & 0.672 & 0.664 & 0.660 & 0.649 & 0.453 & 0.406 & 0.358 & 0.202& 0.620 & 0.598 & 0.300 & 0.294  \\
\rowcolor{gray!30}
Gap (0.947/0.938) &-21.1\%&-20.7\%&-29.0\%&-29.2\%&-30.3\%&\textbf{-30.8\%}&-52.2\%&-56.7\%&-62.2\%&-78.5\%&\textbf{-34.5\%}&\textbf{-36.2\%}&-68.3\%&-68.7\%
 \\
UNIQUE & 0.834 & 0.785 & 0.762 & 0.737 & 0.581 & 0.551 & 0.536 & 0.409 & 0.247 & 0.180 & 0.572 & 0.549 & 0.290 & 0.278 \\
\rowcolor{gray!30}
Gap (0.901/0.896) &-7.4\%&\textbf{-12.4\%}&\textbf{-15.4\%}&\textbf{-17.7\%}&-35.5\%&-38.5\%&-40.5\%&-54.4\%&-72.6\%&-79.9\%&-36.5\%&-38.7\%&-67.8\%&-69.0\%
 \\
CONTRIQUE & 0.722 & 0.720 & 0.315 & 0.183 & 0.636 & 0.546 & 0.749 & 0.674  & 0.128 & 0.091& 0.551 & 0.524 & 0.269 & 0.244 \\
\rowcolor{gray!30}
Gap (0.906/0.894) &-20.3\%&-23.7\%&-65.2\%&-80.6\%&\textbf{-29.8\%}&-42.2\%&-17.3\%&\textbf{-28.6\%}&-85.9\%&-90.4\%&-43.6\%&-44.5\%&-70.3\%&-74.2\%
 \\
MANIQA & 0.735 & 0.696 & 0.604 & 0.542 & 0.397 & 0.293 & 0.362 & 0.073 & 0.208 & 0.146& 0.447 & 0.389 & 0.179 & 0.216  \\
\rowcolor{gray!30}
Gap (0.946/0.944) &-22.3\%&-26.3\%&-36.2\%&-42.6\%&-58.0\%&-69.0\%&-61.7\%&-92.3\%&-78.0\%&-84.5\%&-52.7\%&-58.8\%&-81.1\%&-77.1\%
 \\
VCRNet & 0.708 & 0.690 & 0.622 & 0.592 & 0.544 & 0.517 & 0.317 & 0.286 & 0.121 & 0.120& 0.505 & 0.473 & 0.230 & 0.211  \\
\rowcolor{gray!30}
Gap (0.974/0.973) &-27.3\%&-29.1\%&-36.1\%&-39.2\%&-44.1\%&-46.9\%&-67.5\%&-70.6\%&-87.6\%&-87.7\%&-48.2\%&-51.4\%&-76.4\%&-78.3\%
 \\
LIQE & 0.850 & 0.793 & 0.763 & 0.745 & 0.500 & 0.434 & 0.769 & 0.602 & 0.201 & 0.128& 0.575 & 0.553 & 0.380 & 0.364  \\
\rowcolor{gray!30}
Gap (0.908/0.919)
&\textbf{-6.4\%}&-13.7\%&-16.0\%&-18.9\%&-44.9\%&-52.8\%&\textbf{-15.3\%}&-34.5\%&-77.9\%&-86.1\%&-36.7\%&-39.8\%&\textbf{-58.1\%}&\textbf{-60.4\%}
 \\
QualiCLIP & 0.853 & 0.808 & 0.757 & 0.734 & 0.575 & 0.468 & 0.815 & 0.684 & 0.128 & 0.003& 0.528 & 0.489 & 0.232 & 0.227  \\
\rowcolor{gray!30}
Gap (-/-)
 &- &- &- &- &- &- &- &- &- &- &- &- & -& -\\
Mean Gap &-18.0 \%&-20.7 \%&-33.3 \%&-37.9 \%&-39.4 \%&-44.7 \%&-45.7 \%&-57.2 \%&-75.1 \%&-82.9 \%&-41.1 \%&-43.9 \%&-71.2 \%&-72.1\%
 \\
\bottomrule
\end{tabular}
\label{tab:q1}
\end{table*}

\textbf{$\mathbf{\romannumeral1)}$ Does there exist a gap between BIQA and BPIQA?} As mentioned before, the panoramic image differs substantially from the 2D planar image in many aspects, suggesting that BIQA models are hard to substitute the BPIQA models to assess the perceptual quality of panoramic images. However, we argue that if a PIQA database suffers from the \textit{easy-database} issue, it could narrow the gap between BIQA and BPIQA due to the lack of viewing characteristics of those panoramic images, and vice versa. As shown in Fig.~\ref{fig:analysis_framework} (a), we test a series of pre-trained BIQA models on the PIQA database $\mathcal{D}$ for measuring this gap. Similar to the previous work~\cite{sun2024analysis}, the gap can be calculated by:
\begin{align}
\label{eq:gain}
  Gap=\frac{P_{test}-P_{ori}}{P_{ori}}\times100.
\end{align}
where $P_{ori}$ and $P_{test}$ denote the original and new test performance, respectively.
As a result, a high gap means that the panoramic images in this database are with heterogeneous distortion and the low gap indicates homogeneous distortion. To quantitatively describe this gap in each PIQA database, a test experiment is conducted and the results are shown in Tab.~\ref{tab:q1}, where we have several significant observations. First, the CVIQ database has the smallest mean gap (nearly 0.2 in terms of PLCC and SRCC) especially UNIQUE and LIQE demonstrate transfer performance with only a slight decrease, indicating that the panoramic images in CVIQ are visually similar to 2D planar images. This similarity allows BIQA models to easily transfer their success directly to BPIQA. As a result, CVIQ fails to exhibit the features of panoramic images, leading to an apparent waste of the human labeling budget. Second, these BIQA models exhibit significant performance degradation on the JUFE-10K database, indicating that the heterogeneously distorted panoramic images in JUFE-10K confuse the BIQA models, and successfully create a gap between BIQA and BPIQA models. Besides, it encourages BPIQA models to introduce special designs to deal with this problem rather than simply modifying BIQA models. It is worth noting that the OSIQA database focuses on stitching distortions in panoramic images, arising from complex structural warping that significantly alters visual features and amplifies the gap between 2D planar images and panoramic images. The OSIQA database also considers the unique impact of viewing angles on perceived quality, further widening this gap. In addition, IQA-ODI also presents a significant challenge to these BIQA models, where the primary reason lies in the distortions introduced by four distinct projection patterns, a characteristic unique to panoramic images. Last but not least, from the experimental results on the CVIQ, OIQA, MVAQD and OIQ-10K databases, we can observe that as the types and levels of distortion increase, the challenges posed to BIQA models become greater. This indicates that diverse and abundant distortions contribute to enhancing the overall quality of a database, making it away from the \textit{easy-database} issue. Likewise, we can intuitively assess the severity of the \textit{easy-database} issue in the other four PIQA databases based on the mean gap. Consequently, we straightforwardly rank all PIQA databases as follows (from smallest to largest mean gap): CVIQ, OIQA, MVAQD, OIQ-10K, IQA-ODI, JUFE-10K, and OSIQA.

\textbf{$\mathbf{\romannumeral2)}$ Whether the BPIQA models require those specific designs?} Naively measuring the gap between BIQA and BPIQA based on the transfer performance of BIQA models provides a rough quality assessment of PIQA databases. To explore the inherent factors of the \textit{easy-database} issue, we examine the relationship between model design and distortion samples by devising a family of simple models. The central concept is illustrated in Fig.~\ref{fig:analysis_framework} (b). For the target database $\mathcal{D}$, we first consider two types of models: $\mathcal{M}_1$ and $\mathcal{M}_2$. Here, $\mathcal{M}_1$ refers to the BPIQA models without specific design, which does not consider users' viewing behavior and inter-viewport interaction for panoramic images, while $\mathcal{M}_2$ denotes the BPIQA models with specific design(s) to enhance their ability to handle complex panoramic-specific distortions. Subsequently, we retrain all models and conduct a computational analysis based on their performance from saturation status and model sensitivity. The qualitative results of $\mathcal{M}_1$ and $\mathcal{M}_2$ are shown in Tab.~\ref{tab:q2}, and the detailed discussions are as follows. 

\begin{table*}[t]
\small
\setlength{\tabcolsep}{3.2pt}
\centering
\caption{Comparison of the Results of BPIQA, VQA and BIQA Models on Those PIQA Databases.}
\small
\begin{tabular}{lccccccccccccccc}
\toprule
 & & \multicolumn{2}{c}{CVIQ (7)} & \multicolumn{2}{c}{OIQA (6)} & \multicolumn{2}{c}{MVAQD (4)} & \multicolumn{2}{c}{IQA-ODI (5)} & \multicolumn{2}{c}{OSIQA (3)}  & \multicolumn{2}{c}{OIQ-10K (2)} & \multicolumn{2}{c}{JUFE-10K (1)}\\
\multirow{-2}{*}{Method} & \multirow{-2}{*}{Type} & PLCC & SRCC & PLCC & SRCC & PLCC & SRCC & PLCC & SRCC & PLCC & SRCC& PLCC & SRCC & PLCC & SRCC  \\
\midrule
MC360IQA & $\mathcal{M}_2$ & 0.951 & 0.914 & 0.925 & 0.919 & 0.555 & 0.382 & 0.812 & 0.742 & 0.387 & 0.275 & 0.721 & 0.710 & 0.620 & 0.620\\
Assessor360 & $\mathcal{M}_2$ & \textbf{0.977} & 0.964 & \textbf{0.975} & \textbf{0.980} & \textbf{0.972} & \textbf{0.961} & 0.963 & \textbf{0.957}  & 0.832 & 0.533 & \textbf{0.790} & \textbf{0.773} & 0.694 & 0.690 \\
OIQAND & $\mathcal{M}_2$ & 0.976 & \textbf{0.967} & 0.938 & 0.937 & 0.924 & 0.903 & \textbf{0.965} & 0.927  & \textbf{0.864} & \textbf{0.821} & 0.755 & 0.740 & \textbf{0.800} & 0.800 \\
VSFA & $\mathcal{M}_2$ & 0.965 & 0.953 & 0.963 & 0.961 & 0.924 & 0.932 & 0.940 & 0.930  & 0.801& 0.742 & 0.379 & 0.365 & 0.793 & \textbf{0.810}\\
Trans-VSFA & $\mathcal{M}_2$ & 0.965 & 0.949 & 0.960 & 0.953 & 0.921 & 0.919 & 0.956 & 0.929 & 0.856 & 0.704 & 0.112 & 0.068 & 0.764 & 0.737 \\
FAST-VQA & $\mathcal{M}_2$ & 0.182 & 0.362 & 0.108 & 0.220 & 0.484 & 0.532 & 0.199 & 0.214 & 0.258 & 0.317 & 0.097 & 0.115 & 0.075 & 0.077 \\
\midrule
ResNet-50-P & $\mathcal{M}_1$ & 0.980 & 0.976 & \textbf{0.971} & \textbf{0.969} & 0.897 & 0.890 & 0.963 & 0.930 & \textbf{0.822}
 & 0.675 & 0.808 & 0.814 & \textbf{0.783} & \textbf{0.783} \\
ResNet-152-P & $\mathcal{M}_1$ & 0.977 & 0.974 & 0.967 & 0.967 & 0.858 & 0.854 & 0.957 & 0.910 & 0.730 & 0.597 & 0.810 & 0.814 & 0.781 & 0.782 \\
Swin-v2-T-P & $\mathcal{M}_1$ & 0.974 & 0.964 & 0.968 & 0.967 & 0.907 & 0.899 & 0.965 & 0.929 & 0.818 & 0.727 & 0.797 & 0.797 & 0.737 & 0.730 \\
Swin-v2-B-P & $\mathcal{M}_1$ & 0.976 & 0.968 & 0.955 & 0.953 & 0.924 & 0.908 & \textbf{0.968} & 0.926 & 0.818 & 0.727 & 0.780 & 0.774 & 0.551 & 0.517\\
MaxViT-T-P & $\mathcal{M}_1$ & \textbf{0.984} & \textbf{0.980} & 0.952 & 0.953 & 0.944 & 0.936 & 0.967 & 0.923 & 0.814 & 0.665 & 0.808 & 0.808 & 0.771 & 0.770 \\
HyperIQA & $\mathcal{M}_1$ & 0.973 & 0.960 & 0.970 & 0.968 & 0.822 & 0.817 & 0.929 & 0.887 & 0.717 & \textbf{0.820} & 0.802 & 0.805 & 0.763 & 0.763 \\
MANIQA & $\mathcal{M}_1$ & 0.981 & 0.973 & 0.963 & 0.962 & \textbf{0.965} & \textbf{0.959} & 0.967 & \textbf{0.933} & 0.350 & 0.411 & \textbf{0.821} & \textbf{0.822} & 0.749 & 0.747 \\
VCRNet & $\mathcal{M}_1$ & 0.830 & 0.866 & 0.880 & 0.884 & 0.843 & 0.862 & 0.738 & 0.819 & 0.307 & 0.356 & 0.803 & 0.751 & 0.737 & 0.712 \\
QualiCLIP & $\mathcal{M}_1$ & 0.982 & 0.978 & 0.883 & 0.878 & 0.850 & 0.843 & 0.987 & 0.981 & 0.572 & 0.604 & 0.414 & 0.395 & 0.318 & 0.211 \\
\bottomrule
\end{tabular}
\label{tab:q2}
\end{table*}

\textbf{Saturation status analysis}. For the CVIQ and OIQA databases, most models achieve saturated performance (in terms of 0.96 in terms of PLCC and SRCC), with no significant difference in performance regardless of whether they have specific designs or not. This suggests that the samples in the two databases are relatively simple. In other words, adding fewer types of distortions to panoramic images makes it difficult to provide the model with sufficient learnable information. For MVAQD and IQA-ODI, the saturation condition is less severe and presents challenges for certain models, such as MC360IQA and HyperIQA. This could be attributed to the more primitive feature extractor (\ie, ResNet-50) used in these models, which hampers the degradation modeling of panoramic images due to its limited feature extraction capabilities. Note that MVAQD has one more distortion type than OIQA, while one less distortion level. However, MVAQD poses a greater challenge to models, which suggests that the variety of distortion types contributes more to model's ability to learn distortion features than the number of distortion levels. Surprisingly, the OIQ-10K and JUFE-10K databases reveal the failure of most models, achieving an undersaturated status (nearly 0.8 in terms of PLCC and SRCC), and even causing certain models to fail, meaning that the samples in these two databases are troublesome. Clearly, this issue mainly arises from heterogeneous distortion. OIQ-10K contains both homogeneous and heterogeneous distortions, whereas JUFE-10K consists entirely of heterogeneous distortions. Besides, all three VQA models show less-promising performance on the OIQ-10K database, while two of them perform well on JUFE-10K. This suggests that VQA models may effectively handle temporal information in heterogeneously distorted images (arising from discontinuous viewing experiences) through spatiotemporal modules. However, they struggle when dealing with complex distortion scenarios simultaneously, as OIQ-10K includes a greater variety of distortion types and diverse resolutions. The images in the OSIQA database contain local distortions caused by stitching operations. Due to the small number of samples and the limited variety of distortions, the challenges posed to the model are not as significant as those in the first two databases. Eventually, we successfully demonstrate the \textit{truth} of current status of BPIQA, \ie, there is still room to advance the BPIQA, rather than tend to perfection as revealed in Fig.~\ref{fig:cgp}.

\textbf{Model sensitivity analysis}. A database with the \textit{easy-database} issue results in uniform model performance regardless of design. Conversely, a high-quality database requires specialized designs, making performance sensitive to these variations. Inspired by the previous work~\cite{sun2024analysis}, we explore the \textit{easy-database} issue by quantitatively examining the impact of specific design by comparing with $\mathcal{M}_1$ and $\mathcal{M}_2$. As shown in Tab.~\ref{tab:q2}, the CVIQ demonstrates a similar performance across all models, suggesting that the distortions in this database are relatively less characterized. This is due to CVIQ having only one type of distortion (\textit{i.e.,} compressed distortion) and resolution, and the image content lacks distinctiveness. On the other hand, OIQA exhibits relatively mild symptoms, as it contains a greater variety of distortion types, resolutions, and visual content. Nevertheless, it implies that the basic feature extractor and quality regressor are sufficient to handle the simple distortions (\textit{i.e.}, few distortion types and levels) in the two databases. Due to differences in handling one of the distortion types on the CMP and the varying distortion situations in MVAQD, models need specialized designs to address this panoramic-specific distortion, which increases the model sensitivity. For example, MC360IQA predicts the perceptual quality of panoramic images by utilizing a FC to regress the features of six viewport images into a single score, \ie, the integrated mapping with basic operation. However, it neglects proper interaction modeling to address CMP-based distortion, resulting in severe performance degradation. Besides, we find that transformer-based and hybrid-based models are superior to CNN-based models as the feature extractor in this case, indicating the ability of global modeling in transformer is of great importance to address panoramic-specific distortion. Like MVAQD, samples in IQA-ODI have distortions added across multiple projection patterns, which facilitates the form of panoramic-specific distortion. However, the single distortion type limits the model's ability to learn more about distortion, leading to similar results. For the OIQ-10K, it is obvious that the performance of $\mathcal{M}_1$ is better than $\mathcal{M}_2$, a plausible explanation may be that the samples in OIQ-10K have both \textit{homogeneous distortion} and \textit{heterogeneous distortion}, which imposes higher demands on model design. However, inappropriate special designs can instead disrupt model predictions, revealing that there are still no reasonable models fundamentally capable of addressing both homogeneous and heterogeneous distortions simultaneously. JUFE-10K only contains samples with \textit{heterogeneous distortion}, which compels models concentrate on proposing more specific designs to address the common discontinuity of experience (caused by local distortion) that is closed to temporal modeling problem. From the Tab.~\ref{tab:q2}, we surprisingly observe that the VSFA (\ie, a VQA model) obtains the best results on JUFE-10K, revealing that VQA models could handle the heterogeneous distortion in BPIQA. In addition, we also find that QualiCLIP achieves worse results on heterogeneous distortion (\textit{i.e.}, OSIQA, OIQ-10K and JUFE-10K) than homogeneous distortion. The possible reason is that the textual features interfere with the model's ability to learn heterogeneous distortions. On the contrary, OIQAND demonstrates promising performance on the three databases, which can be attributed to its specific viewport interaction approach. It enables OIQAND to adaptively adjust the weights of each viewport through the Viewports Vit module. Note that FAST-VQA performs poorly on most databases. This could be due to the sampling method in the model, which significantly reduces the visual features, thereby failing to provide sufficient information for training. Additionally, the quality regressor consists of 3D convolutional layer and the global average pooling (GAP) may not be suitable for BPIQA. Finally, we empirically rank all PIQA databases according to the saturation status (from mild to severe): JUFE-10K, OIQ-10K, OSIQA, MVAQD, IQA-ODI, OIQA, and CVIQ.

\textbf{$\mathbf{\romannumeral3)}$ Does it empower BPIQA to deal with hard samples?} A BPIQA model with outstanding generalization ability accurately assesses the perceptual quality of various in-the-wild panoramic images. However, due to its limited sample coverage, the PIQA database with \textit{easy-database} issue cannot empower BPIQA models to handle hard samples in the real world and vice versa. From this perspective, as shown in Fig.~\ref{fig:analysis_framework} (c), we conduct a series of cross-validation experiments to explore the \textit{easy-database} issue. Specifically, similar to work~\cite{sun2024analysis}, we quantitatively measure the ability of the target database to enhance the model's generalization performance by calculating the gain between the performance on the training database and the performance on the testing database. The calculation of gain is the same as that of gap. Lower gain indicates more serious sample issues within the database.

\textbf{CVIQ}. The results are presented in Tab.~\ref{tab:q3_cviq}. We observe that the performance of most BPIQA models dramatically drops when tested on other databases, suggesting that CVIQ may not fully equip models to handle challenging samples in different domains. However, it is worth noting that the results of testing on IQA-ODI are relatively impressive. This could be attributed to the CVIQ focus on compressed distortion in panoramic images, while IQA-ODI happens to only contain JPEG compression. This indicates that CVIQ does enable BPIQA models to handle samples with compressed distortion positively.

\begin{table*}[t]
\centering
\caption{Cross-database Evaluation. The BPIQA Models are Trained on CVIQ and Tested on Other Six Databases.}
\small
\begin{tabular}{lcccccccccccccc}
\toprule
 & \multicolumn{2}{c}{OIQA} & \multicolumn{2}{c}{MVAQD} & \multicolumn{2}{c}{IQA-ODI} & \multicolumn{2}{c}{OSIQA}& \multicolumn{2}{c}{OIQ-10K} & \multicolumn{2}{c}{JUFE-10K} \\
\multirow{-2}{*}{Method} & PLCC & SRCC & PLCC & SRCC & PLCC & SRCC & PLCC & SRCC& PLCC & SRCC & PLCC & SRCC \\
\midrule
MC360IQA & 0.424 & 0.386 & 0.229 & 0.163 & 0.812 & 0.658 & 0.106 & 0.132 & 0.347 & 0.279 & 0.031 & 0.028 \\
\rowcolor{gray!30}
Gain & -55.4\%&-57.8\%&-75.9\%&-82.2\%&-14.6\%&-28.0\%&-88.9\%&-85.6\%&-63.5\%&-69.5\%&-96.7\%&-96.9\%
 \\
Assessor360 & 0.542 & 0.463 & 0.629 & 0.564 & 0.905 & 0.860 & 0.171 & 0.125 & 0.430 & 0.336 & 0.061 & 0.249 \\
\rowcolor{gray!30}
Gain & -44.5\%&-52.0\%&-35.6\%&-41.5\%&-7.4\%&-10.8\%&-82.5\%&-87.0\%&-56.0\%&-65.1\%&-92.7\%&-93.7\%
 \\
OIQAND & 0.636 & 0.615 & 0.605 & 0.494 & 0.776 & 0.562 & 0.214 & 0.151 & 0.296 & 0.142 & 0.013 & 0.002 \\
\rowcolor{gray!30}
Gain & -34.8\%&-36.4\%&-38.0\%&-48.9\%&-20.5\%&-41.9\%&-78.1\%&-84.4\%&-69.7\%&-85.3\%&-98.7\%&-99.8\%
 \\
\midrule
ResNet-50-P & 0.679 & 0.662 & 0.385 & 0.356 & 0.824 & 0.712 & 0.225 & 0.292 & 0.327 & 0.250 & 0.064 & 0.051 \\
\rowcolor{gray!30}
Gain & -30.7\%&-32.2\%&-60.7\%&-63.5\%&-15.9\%&-27.0\%&-77.0\%&-70.1\%&-66.6\%&-74.4\%&-93.5\%&-94.8\%
\\
ResNet-152-P & 0.702 & 0.685 & 0.577 & 0.553 & 0.835 & 0.765 & 0.142 & 0.144 & 0.397 & 0.262 & 0.054 & 0.046 \\
\rowcolor{gray!30}
Gain & -28.1\%&-29.7\%&-40.9\%&-43.2\%&-14.5\%&-21.5\%&-85.5\%&-85.2\%&-59.4\%&-73.1\%&-94.5\%&-95.3\%
\\
Swin-v2-T-P & 0.792 & 0.780 & 0.792 & 0.736 & 0.878 & 0.796 & 0.013 & 0.025 & 0.438 & 0.331 & 0.167 & 0.149 \\
\rowcolor{gray!30}
Gain & -18.7\%&-19.1\%&-18.7\%&-23.7\%&-9.9\%&-17.4\%&-98.7\%&-97.4\%&-55.0\%&-65.7\%&-82.9\%&-84.5\%
 \\
Swin-v2-B-P & 0.777 & 0.773 & 0.526 & 0.456 & 0.836 & 0.682 & 0.268 & 0.223 & 0.443 & 0.371 & 0.208 & 0.198 \\
\rowcolor{gray!30}
Gain & -20.4\%&-20.1\%&-46.1\%&-52.9\%&-14.3\%&-29.5\%&-72.5\%&-77.0\%&-54.6\%&-61.7\%&-78.7\%&-79.5\%
 \\
MaxViT-T-P & 0.636 & 0.603 & 0.582 & 0.553 & 0.893 & 0.836  & 0.234 & 0.281 & 0.401 & 0.230 & 0.053 & 0.048\\
\rowcolor{gray!30}
Gain & -35.4\%&-38.5\%&-40.9\%&-43.6\%&-9.2\%&-14.7\%&-76.2\%&-71.3\%&-59.2\%&-76.5\%&-94.6\%&-95.1\%
 \\
 Mean Gain & -33.5\%&-35.7\%&-44.6\%&-49.9\%&-13.3\%&-23.9\%&-82.4\%&-82.2\%&-60.5\%&-71.4\%&-91.5\%&-92.5\%
\\
\bottomrule
\end{tabular}
\label{tab:q3_cviq}
\end{table*}

\textbf{OIQA}. As shown in Tab.~\ref{tab:q3_oiqa}, the gain from testing on other databases increases when models are trained on OIQA compared to CVIQ, especially most models obtaining considerable generalization performance on the CVIQ. That is to say, OIQA empowers BPIQA models to effectively deal with the samples in CVIQ and yields more gain in the training process than CVIQ. This might be due to OIQA containing various distortion situations and plentiful content, despite the number of samples being less than CVIQ.

\begin{table*}[t]
\centering
\caption{Cross-database Evaluation. The BPIQA Models are Trained on OIQA and Tested on Other Six Databases.}
\small
\begin{tabular}{lcccccccccccc}
\toprule
 & \multicolumn{2}{c}{CVIQ} & \multicolumn{2}{c}{MVAQD} & \multicolumn{2}{c}{IQA-ODI} & \multicolumn{2}{c}{OSIQA}& \multicolumn{2}{c}{OIQ-10K} & \multicolumn{2}{c}{JUFE-10K} \\
\multirow{-2}{*}{Method} & PLCC & SRCC & PLCC & SRCC & PLCC & SRCC & PLCC & SRCC& PLCC & SRCC & PLCC & SRCC \\
\midrule
MC360IQA & 0.868 & 0.853 & 0.537 & 0.492 & 0.764 & 0.638  & 0.054 & 0.033 & 0.565 & 0.616 & 0.184 & 0.179\\
\rowcolor{gray!30}
Gain & -6.2\%&-7.2\%&-41.9\%&-46.5\%&-17.4\%&-30.6\%&-94.2\%&-96.4\%&-38.9\%&-43.9\%&-80.1\%&-80.5\%\\
Assessor360 & 0.893 & 0.853 & 0.660 & 0.642 & 0.915 & 0.873  & 0.242 & 0.325 & 0.510 & 0.348 & 0.041 & 0.034\\
\rowcolor{gray!30}
Gain & -8.4\%&-13.0\%&-32.3\%&-34.5\%&-6.2\%&-10.9\%&-75.2\%&-66.8\%&-47.7\%&-64.5\%&-95.8\%&-96.5
\% \\
OIQAND & 0.915 & 0.905 & 0.797 & 0.747 & 0.850 & 0.740  & 0.132 & 0.121 & 0.580 & 0.520 & 0.198 & 0.185\\
\rowcolor{gray!30}
Gain & -2.5\%&-3.4\%&-15.0\%&-20.3\%&-9.4\%&-21.0\%&-85.9\%&-87.1\%&-38.2\%&-44.5\%&-78.9\%&-80.3\%
 \\
\midrule
ResNet-50-P & 0.921 & 0.901 & 0.700 & 0.651 & 0.788 & 0.621 & 0.129 & 0.070 & 0.621 & 0.489 & 0.164 & 0.146 \\
\rowcolor{gray!30}
Gain & -5.1\%&-7.0\%&-27.9\%&-32.8\%&-18.8\%&-35.9\%&-86.7\%&-92.8\%&-43.5\%&-49.5\%&-83.1\%&-84.9
\%\\
ResNet-152-P & 0.938 & 0.931 & 0.708 & 0.695 & 0.823 & 0.710  & 0.226 & 0.060 & 0.580 & 0.535 & 0.210 & 0.196\\
\rowcolor{gray!30}
Gain & -3.0\%&-3.7\%&-26.8\%&-28.1\%&-14.9\%&-26.6\%&-76.6\%&-93.8\%&-40.0\%&-44.7\%&-78.3\%&-79.7
 \\
Swin-v2-T-P & 0.931 & 0.916 & 0.821 & 0.804 & 0.851 & 0.749  & 0.225 & 0.181 & 0.573 & 0.519 & 0.227 & 0.187\\
\rowcolor{gray!30}
Gain & -3.8\%&-5.3\%&-15.2\%&-16.9\%&-12.1\%&-22.5\%&-76.8\%&-81.3\%&-40.8\%&-46.3\%&-76.5\%&-80.7
\%\\
Swin-v2-B-P & 0.934 & 0.913 & 0.840 & 0.803 & 0.865 & 0.768 & 0.196 & 0.074 & 0.584 & 0.523 & 0.216 & 0.168 \\
\rowcolor{gray!30}
Gain & -2.2\%&-4.2\%&-12.0\%&-15.7\%&-9.4\%&-19.4\%&-79.5\%&-92.2\%&-38.8\%&-45.1\%&-77.4\%&-82.4
\% \\
MaxViT-T-P & 0.922 & 0.902 & 0.663 & 0.657 & 0.810 & 0.718  & 0.144 & 0.096 & 0.559 & 0.505 & 0.107 & 0.097\\
\rowcolor{gray!30}
Gain & -3.2\%&-5.4\%&-30.4\%&-31.1\%&-14.9\%&-24.7\%&-84.9\%&-89.9\%&-41.3\%&-47.0\%&-88.8\%&-89.8
\% \\
 Mean Gain & -4.3\%&-6.1\%&-25.2\%&-28.2\%&-12.9\%&-24.0\%&-82.5\%&-87.5\%&-41.1\%&-48.2\%&-82.4\%&-84.4\%
\\
\bottomrule
\end{tabular}
\label{tab:q3_oiqa}
\end{table*}

\textbf{MVAQD}. We find that some BPIQA models achieve acceptable gains on CVIQ, OIQA, and IQA-ODI as seen in Tab.~\ref{tab:q3_mvaqd}, such as the CNN-based ResNet-50-P, Transformer-based Swin-v2-T-P and its variant Swin-v2-B-P. One reasonable explanation is that MVAQD includes a rich array of distortion scenarios and resolutions, which is more consistent with the distorted panoramic images in the real world. Although MVAQD enables BPIQA models to handle samples with \textit{homogeneous distortion}, it does not empower models to deal with samples exhibiting \textit{heterogeneous distortion}, as evidenced by the results on OIQ-10K and JUFE-10K.

\begin{table*}[t]
\centering
\small
\caption{Cross-database Evaluation. The BPIQA Models are Trained on MVAQD and Tested on Other Six Databases.}
\begin{tabular}{lcccccccccccc}
\toprule
 & \multicolumn{2}{c}{CVIQ} & \multicolumn{2}{c}{OIQA} & \multicolumn{2}{c}{IQA-ODI} & \multicolumn{2}{c}{OSIQA} & \multicolumn{2}{c}{OIQ-10K} & \multicolumn{2}{c}{JUFE-10K} \\
\multirow{-2}{*}{Method} & PLCC & SRCC & PLCC & SRCC & PLCC & SRCC & PLCC & SRCC& PLCC & SRCC & PLCC & SRCC \\
\midrule
MC360IQA & 0.116 & 0.123 & 0.131 & 0.127 & 0.102 & 0.114  & 0.032 & 0.018 & 0.119 & 0.112 & 0.097 & 0.085\\
\rowcolor{gray!30}
Gain & -79.1\%&-67.8\%&-76.4\%&-66.8\%&-81.6\%&-70.2\%&-94.2\%&-95.3\%&-78.6\%&-70.7\%&-82.5\%&-77.7\%\\
Assessor360 & 0.855 & 0.885 & 0.785 & 0.778 & 0.868 & 0.862 & 0.129 & 0.165 & 0.499 & 0.377& 0.064 & 0.638 \\
\rowcolor{gray!30}
Gain & -12.0\%&-7.9\%&-19.2\%&-19.0\%&-10.7\%&-10.3\%&-86.7\%&-82.8\%&-48.7\%&-60.8\%&-94.0\%&-93.3\%
 \\
OIQAND & 0.673 & 0.728 & 0.834 & 0.822 & 0.833 & 0.823 & 0.058 & 0.027 & 0.521 & 0.410 & 0.048 & 0.030 \\
\rowcolor{gray!30}
Gain & -27.2\%&-19.4\%&-9.7\%&-9.0\%&-9.8\%&-8.9\%&-93.7\%&-97.0\%&-43.6\%&-54.6\%&-94.8\%&-96.7\%
 \\
\midrule
ResNet-50-P & 0.871 & 0.852 & 0.873 & 0.861 & 0.855 & 0.791 & 0.208 & 0.114 & 0.515 & 0.405 & 0.047 & 0.045 \\
\rowcolor{gray!30}
Gain &-2.9\%&-4.3\%&-2.7\%&-3.3\%&-4.7\%&-11.1\%&-76.8\%&-87.2\%&-42.6\%&-54.5\%&-94.8\%&-94.9\%
  \\
ResNet-152-P & 0.731 & 0.750 & 0.694 & 0.681 & 0.784 & 0.728 & 0.201 & 0.199 & 0.500 & 0.370 & 0.047 & 0.018 \\
\rowcolor{gray!30}
Gain & -14.8\%&-12.2\%&-19.1\%&-20.3\%&-8.6\%&-14.8\%&-76.6\%&-76.7\%&-41.7\%&-56.7\%&-94.5\%&-97.9\%
 \\
Swin-v2-T-P & 0.792 & 0.835 & 0.854 & 0.845 & 0.874 & 0.809 & 0.090 & 0.043 & 0.512 & 0.399 & 0.052 & 0.033 \\
\rowcolor{gray!30}
Gain & -12.7\%&-7.1\%&-5.8\%&-6.0\%&-3.6\%&-10.0\%&-90.1\%&-95.2\%&-43.6\%&-55.6\%&-94.3\%&-96.3\%
 \\
Swin-v2-B-P & 0.887 & 0.911 & 0.909 & 0.901 & 0.860 & 0.780 & 0.040 & 0.160 & 0.520 & 0.376 & 0.068 & 0.071 \\
\rowcolor{gray!30}
Gain & -4.0\%&+0.3\%&-1.6\%&-0.8\%&-6.9\%&-14.1\%&-95.7\%&-82.4\%&-43.7\%&-58.6\%&-92.6\%&-92.2\%
 \\
MaxViT-T-P & 0.762 & 0.788 & 0.781 & 0.741 & 0.837 & 0.734 & 0.340 & 0.325 & 0.525 & 0.449 & 0.033 & 0.031 \\
\rowcolor{gray!30}
Gain &-19.3\%&-15.8\%&-17.3\%&-20.8\%&-11.3\%&-21.6\%&-64.0\%&-65.3\%&-44.4\%&-52.0\%&-96.5\%&-96.7\%
 \\
  Mean Gain & -21.5\%&-16.8\%&-19.0\%&-18.2\%&-17.2\%&-20.1\%&-84.7\%&-85.2\%&-48.4\%&-57.9\%&-93.0\%&-93.2\%
\\
\bottomrule
\end{tabular}
\label{tab:q3_mvaqd}
\end{table*}

\textbf{IQA-ODI}. The results are shown in Tab.~\ref{tab:q3_iqa_odi}, even though IQA-ODI contains a substantial number of samples, the single distortion situation limits BPIQA models to learn more useful information of distortion, leading to an unsatisfactory gain on other databases. Another plausible reason is that IQA-ODI focuses on studying the impact of map projection, which deviates somewhat from the distortion domain.

\begin{table*}[t]
\centering
\small
\caption{Cross-database Evaluation. The BPIQA Models are Trained on IQA-ODI and Tested on Other Six Databases.}
\begin{tabular}{lcccccccccccc}
\toprule
\multirow{2}{*}{Method} & \multicolumn{2}{c}{CVIQ} & \multicolumn{2}{c}{OIQA} & \multicolumn{2}{c}{MVAQD} & \multicolumn{2}{c}{OSIQA}& \multicolumn{2}{c}{OIQ-10K} & \multicolumn{2}{c}{JUFE-10K} \\
 & PLCC & SRCC & PLCC & SRCC & PLCC & SRCC & PLCC & SRCC& PLCC & SRCC & PLCC & SRCC  \\
\midrule
MC360IQA & 0.741 & 0.676 & 0.557 & 0.543 & 0.722 & 0.640 & 0.174 & 0.289 & 0.420 & 0.380 & 0.078 & 0.068 \\
\rowcolor{gray!30}
Gain & -8.7\%&-8.9\%&-31.4\%&-26.8\%&-11.1\%&-13.7\%&-78.6\%&-61.1\%&-48.3\%&-48.8\%&-90.4\%&-90.8\%
 \\
Assessor360 & 0.767 & 0.726 & 0.380 & 0.110 & 0.541 & 0.476  & 0.129 & 0.165 & 0.356 & 0.177 & 0.100 & 0.089\\
\rowcolor{gray!30}
Gain &-20.4\%&-24.1\%&-60.5\%&-88.5\%&-43.8\%&-50.3\%&-86.6\%&-82.8\%&-63.0\%&-81.5\%&-89.6\%&-90.7\%
\\
OIQAND & 0.596 & 0.591 & 0.377 & 0.083 & 0.408 & 0.152 & 0.217 & 0.190 & 0.341 & 0.138& 0.095 & 0.089\\
\rowcolor{gray!30}
Gain & -38.2\%&-36.2\%&-60.9\%&-91.0\%&-57.7\%&-83.6\%&-77.5\%&-79.5\%&-64.7\%&-85.1\%&-90.2\%&-90.4\%
\\
\midrule
ResNet-50-P & 0.840 & 0.825 & 0.381 & 0.273 & 0.551 & 0.548 & 0.130 & 0.055 & 0.375 & 0.330 & 0.048 & -0.008 \\
\rowcolor{gray!30}
Gain & -12.8\%&-11.3\%&-60.4\%&-70.6\%&-42.8\%&-41.1\%&-86.5\%&-94.1\%&-61.1\%&-64.5\%&-95.0\%&-100.9\%
 \\
ResNet-152-P & 0.809 & 0.783 & 0.375 & 0.191 & 0.603 & 0.570  & 0.222 & 0.229 & 0.383 & 0.310 & 0.045 & 0.022\\
\rowcolor{gray!30}
Gain & -15.5\%&-14.0\%&-60.8\%&-79.0\%&-37.0\%&-37.4\%&-76.8\%&-74.8\%&-60.0\%&-65.9\%&-95.3\%&-97.6\%
 \\
Swin-v2-T-P & 0.859 & 0.838 & 0.402 & 0.366 & 0.640 & 0.602  & 0.243 & 0.231 & 0.449 & 0.368 & 0.055 & 0.050\\
\rowcolor{gray!30}
Gain &-11.0\%&-9.8\%&-58.3\%&-60.6\%&-33.7\%&-35.2\%&-74.8\%&-75.1\%&-53.5\%&-60.4\%&-94.3\%&-94.6\%
 \\
Swin-v2-B-P & 0.779 & 0.717 & 0.369 & 0.259 & 0.470 & 0.403 & 0.129 & 0.060 & 0.359 & 0.267 & 0.025 & 0.017 \\
\rowcolor{gray!30}
Gain & -19.5\%&-22.6\%&-61.9\%&-72.0\%&-51.4\%&-56.5\%&-86.7\%&-93.5\%&-62.9\%&-71.2\%&-97.4\%&-98.2\%
 \\
MaxViT-T-P & 0.847 & 0.832 & 0.385 & 0.265 & 0.499 & 0.336 & 0.175 & 0.216 & 0.400 & 0.291 & 0.042 & 0.008 \\
\rowcolor{gray!30}
Gain & -12.4\%&-9.9\%&-60.2\%&-71.3\%&-48.4\%&-63.6\%&-81.9\%&-76.6\%&-58.6\%&-68.5\%&-95.7\%&-99.1\%\\ 
  Mean Gain & -17.3\%&-17.1\%&-56.8\%&-70.0\%&-40.7\%&-47.7\%&-81.2\%&-79.7\%&-59.0\%&-68.2\%&-93.5\%&-95.3\%
\\
\bottomrule
\end{tabular}
\label{tab:q3_iqa_odi}
\end{table*}

\textbf{OSIQA}. From the Tab.~\ref{tab:q3_osiqa}, we can clearly observe a significant decline in the generalization performance of nearly all models, indicating that the distortion information provided by OSIQA does not help the models learn the distortion features of panoramic images effectively. The reason lies in the small number of samples and the unique attribute of the distortion (stitching distortion in panoramic images), which makes it difficult to generalize to other types of distortion.

\begin{table*}[t]
\centering
\small
\caption{Cross-database Evaluation. The BPIQA Models are Trained on OSIQA and Tested on Other Six Databases.}
\begin{tabular}{lcccccccccccc}
\toprule
\multirow{2}{*}{Method} & \multicolumn{2}{c}{CVIQ} & \multicolumn{2}{c}{OIQA} & \multicolumn{2}{c}{MVAQD} & \multicolumn{2}{c}{IQA-ODI}& \multicolumn{2}{c}{OIQ-10K} & \multicolumn{2}{c}{JUFE-10K} \\
 & PLCC & SRCC & PLCC & SRCC & PLCC & SRCC & PLCC & SRCC& PLCC & SRCC & PLCC & SRCC  \\
\midrule
MC360IQA &0.147&0.163&0.148&0.133&0.097&0.190&0.144&0.113&0.119&0.146&0.043&0.043\\
\rowcolor{gray!30}
Gain & -62.0\%&-40.7\%&-61.8\%&-51.6\%&-74.9\%&-30.9\%&-62.8\%&-58.9\%&-69.3\%&-46.9\%&-88.9\%&-84.4\%
 \\
Assessor360 & 0.008&0.309&0.334&0.333&0.056&0.246&0.091&0.059&0.143&0.162&0.040&0.040 \\
\rowcolor{gray!30}
Gain & -70.9\%&-55.9\%&-83.4\%&-78.8\%&-78.7\%&-65.1\%&-87.1\%&-75.0\%&-94.8\%&-99.2\%&-94.5\%&-91.2\%
\\
OIQAND & 0.242&0.235&0.138&0.113&0.177&0.186&0.107&0.133&0.043&0.004&0.046&0.047 \\
\rowcolor{gray!30}
Gain & -72.0\%&-71.4\%&-84.0\%&-86.2\%&-79.5\%&-77.3\%&-87.6\%&-83.8\%&-95.0\%&-99.5\%&-94.7\%&-94.3\%
 \\
\midrule
ResNet-50-P & 0.369&0.136&0.293&0.291&0.166&0.168&0.116&0.001&0.12&0.096&0.004&0.005
 \\
\rowcolor{gray!30}
Gain & -55.1\%&-79.9\%&-64.4\%&-56.9\%&-79.8\%&-75.1\%&-85.9\%&-99.9\%&-85.4\%&-85.8\%&-99.5\%&-99.3\%
 \\
ResNet-152-P & 0.115&0.111&0.154&0.034&0.097&0.094&0.141&0.131&0.05&0.031&0.049&0.037
 \\
\rowcolor{gray!30}
Gain & -84.2\%&-81.4\%&-78.9\%&-94.3\%&-86.7\%&-84.3\%&-80.7\%&-78.1\%&-93.2\%&-94.8\%&-93.3\%&-93.8\%
 \\
Swin-v2-T-P & 0.266&0.249&0.18&0.17&0.214&0.226&0.213&0.23&0.027&0.016&0.098&0.077
 \\
\rowcolor{gray!30}
Gain & -67.5\%&-65.7\%&-78.0\%&-76.6\%&-73.8\%&-68.9\%&-74.0\%&-68.4\%&-96.7\%&-97.8\%&-88.0\%&-89.4\%
 \\
Swin-v2-B-P & 0.133&0.117&0.177&0.175&0.166&0.166&0.14&0.146&0.037&0.025&0.077&0.071
 \\
\rowcolor{gray!30}
Gain & -83.7\%&-83.9\%&-78.4\%&-75.9\%&-79.7\%&-77.2\%&-82.9\%&-79.9\%&-95.5\%&-96.6\%&-90.6\%&-90.2\%
 \\
MaxViT-T-P & 0.191&0.04&0.197&0.195&0.164&0.111&0.125&0.133&0.077&0.067&0.096&0.089
 \\
\rowcolor{gray!30}
Gain & -76.5\%&-94.0\%&-75.8\%&-70.7\%&-79.9\%&-83.3\%&-84.6\%&-80.0\%&-90.5\%&-89.9\%&-88.2\%&-86.6\%
 \\
  Mean Gain & -71.5\%&-71.6\%&-75.6\%&-73.9\%&-79.1\%&-70.3\%&-80.7\%&-78.0\%&-90.0\%&-88.8\%&-92.2\%&-91.1\%
\\
\bottomrule
\end{tabular}
\label{tab:q3_osiqa}
\end{table*}

\textbf{OIQ-10K}. We surprisingly observe that a large portion of BPIQA models trained on the OIQ-10K has a positive gain on other databases as exhibited in Tab.~\ref{tab:q3_oiq_10k}, and the performance of these models merely has slight declines on the JUFE-10K, meaning that it empowers BPIQA models to deal with hard samples with both \textit{homogeneous distortion} and \textit{heterogeneous distortion}. OIQ-10K achieves this level of success due to its samples covering a wide range of panoramic space, including diverse scenes, image resolutions, and types of degradation.

\begin{table*}[t]
\centering
\small
\caption{Cross-database Evaluation. The BPIQA Models are Trained on OIQ-10K and Tested on Other Six Databases.}
\begin{tabular}{lcccccccccccc}
\toprule
\multirow{2}{*}{Method} & \multicolumn{2}{c}{CVIQ} & \multicolumn{2}{c}{OIQA} & \multicolumn{2}{c}{MVAQD} & \multicolumn{2}{c}{IQA-ODI} & \multicolumn{2}{c}{OSIQA} & \multicolumn{2}{c}{JUFE-10K} \\
 & PLCC & SRCC & PLCC & SRCC & PLCC & SRCC & PLCC & SRCC & PLCC & SRCC & PLCC & SRCC\\
\midrule
MC360IQA & 0.705 & 0.740 & 0.818 & 0.809 & 0.643 & 0.533 & 0.496 & 0.436 & 0.013 & 0.008 & 0.319 & 0.253 \\
\rowcolor{gray!30}
Gain & -2.2\%&+4.2\%&+13.5\%&+13.9\%&-10.8\%&-24.9\%&-31.2\%&-38.6\%&-98.2\%&-98.9\%&-55.8\%&-64.4\%
 \\
Assessor360 & 0.842 & 0.869 & 0.909 & 0.899 & 0.710 & 0.680 & 0.801 & 0.628 & 0.149 & 0.186 & 0.624 & 0.613 \\
\rowcolor{gray!30}
Gain & +6.6\%&+12.4\%&+15.1\%&+16.3\%&-10.1\%&-12.0\%&+1.4\%&-18.8\%&-81.1\%&-75.9\%&-21.0\%&-20.7\%
 \\
OIQAND & 0.687 & 0.735 & 0.847 & 0.843 & 0.588 & 0.544 & 0.507 & 0.479 & 0.133
 & 0.061 &0.324 & 0.297 \\
\rowcolor{gray!30}
Gain & -9.0\%&-0.7\%&+12.2\%&+13.9\%&-22.1\%&-26.5\%&-32.8\%&-35.3\%&-82.4\%&-91.8\%&-57.1\%&-59.9\%
 \\
\midrule
ResNet-50-P & 0.671 & 0.692 & 0.884 & 0.872 & 0.776 & 0.736 & 0.772 & 0.630 & 0.179 & 0.161 &0.703 & 0.695 \\
\rowcolor{gray!30}
Gain & -17.0\%&-15.0\%&+9.4\%&+7.1\%&-4.0\%&-9.6\%&-4.5\%&-22.6\%&-77.8\%&-80.2\%&-13.0&-14.6\%
 \\
ResNet-152-P & 0.792 & 0.795 & 0.882 & 0.871 & 0.777 & 0.754 & 0.788 & 0.650 & 0.021 & 0.046 &0.649 & 0.634 \\
\rowcolor{gray!30}
Gain & -2.2\%&-2.3\%&+8.9\%&+7.0\%&-4.1\%&-7.4\%&-2.7\%&-20.1\%&-97.4\%&-94.3\%&-19.9\%&-22.1\%
 \\
Swin-v2-T-P & 0.865 & 0.869 & 0.903 & 0.894 & 0.766 & 0.733 & 0.802 & 0.726 & 0.070 & 0.010 &0.630 & 0.611 \\
\rowcolor{gray!30}
Gain & +8.5\%&+9.0\%&+13.3\%&+12.2\%&-3.9\%&-8.0\%&+0.6\%&-8.9\%&-91.2\%&-98.7\%&-21.0\%&-23.3\%
 \\
Swin-v2-B-P & 0.798 & 0.821 & 0.924 & 0.917 & 0.782 & 0.753 & 0.772 & 0.692 & 0.186 & 0.122 &0.551 & 0.517 \\
\rowcolor{gray!30}
Gain & +2.3\%&+6.1\%&+18.5\%&+18.5\%&+0.3\%&-2.7\%&-1.0\%&-10.6\%&-76.2\%&-84.2\%&-29.4\%&-33.2\%
 \\
MaxViT-T-P & 0.843 & 0.857 & 0.896 & 0.887 & 0.762 & 0.741 & 0.789 & 0.694 & 0.057 & 0.009 &0.681 & 0.669 \\
\rowcolor{gray!30}
Gain & +4.3\%&+6.1\%&+10.9\%&+9.8\%&-5.7\%&-8.3\%&-2.4\%&-14.1\%&-92.9\%&-98.9\%&-15.7\%&-17.2\%
 \\ 
  Mean Gain & -1.1\%&+2.5\%&+12.7\%&+12.3\%&-7.6\%&-12.4\%&-9.1\%&-21.1\%&-87.2\%&-90.4\%&-29.1\%&-31.9\%
\\
\bottomrule
\end{tabular}
\label{tab:q3_oiq_10k}
\end{table*}

\textbf{JUFE-10K}. The results are illustrated in Tab.~\ref{tab:q3_jufe_10k}. 
Despite JUFE-10K containing samples comparable to those in OIQ-10K, the gain on other databases is not entirely satisfactory. A possible explanation is that JUFE-10K focuses on samples with \textit{heterogeneous distortion}, however, so far no BPIQA model has demonstrated the capability to learn and effectively apply local distortion patterns, which could advance the \textit{truth} of the current status of BPIQA by encouraging the emergence of models with specialized designs to address local distortions.

\begin{table*}[t]
\centering
\small
\caption{Cross-database Evaluation. The BPIQA Models are Trained on JUFE-10K and Tested on Other Six Databases.}
\begin{tabular}{lcccccccccccc}
\toprule
\multirow{2}{*}{Method} & \multicolumn{2}{c}{CVIQ} & \multicolumn{2}{c}{OIQA} & \multicolumn{2}{c}{MVAQD} & \multicolumn{2}{c}{IQA-ODI} & \multicolumn{2}{c}{OSIQA} & \multicolumn{2}{c}{OIQ-10K} \\
 & PLCC & SRCC & PLCC & SRCC & PLCC & SRCC & PLCC & SRCC & PLCC & SRCC & PLCC & SRCC\\
\midrule
MC360IQA & 0.456 & 0.438 & 0.471 & 0.390 & 0.320 & 0.096 & 0.117 & 0.117 & 0.168 & 0.136 & 0.416 & 0.409 \\
\rowcolor{gray!30}
Gain & -26.5\%&-28.3\%&-24.0\%&-36.2\%&-48.4\%&-84.3\%&-81.1\%&-80.9\%&-72.9\%&-77.7\%&-32.9\%&-33.1\%
 \\
Assessor360 & 0.423 & 0.478 & 0.243 & 0.240 & 0.363 & 0.182 & 0.089 & 0.089 & 0.252 & 0.281 & 0.368 & 0.376 \\
\rowcolor{gray!30}
Gain & -39.0\%&-30.7\%&-65.0\%&-65.2\%&-47.7\%&-73.6\%&-87.2\%&-87.1\%&-63.7\%&-59.3\%&-47.0\%&-45.5\%
 \\
OIQAND & 0.667 & 0.673 & 0.496 & 0.487 & 0.562 & 0.503 & 0.209 & 0.145 & 0.205 & 0.077 & 0.555 & 0.560 \\
\rowcolor{gray!30}
Gain & -16.6\%&-15.9\%&-38.0\%&-39.1\%&-29.8\%&-37.1\%&-73.9\%&-81.9\%&-74.4\%&-90.4\%&-30.6\%&-30.0\%
 \\
\midrule
ResNet-50-P & 0.506 & 0.543 & 0.516 & 0.448 & 0.512 & 0.408 & 0.182 & 0.080 & 0.164 & 0.054 & 0.504 & 0.507 \\
\rowcolor{gray!30}
Gain &-35.4\%&-30.7\%&-34.1\%&-42.8\%&-34.6\%&-47.9\%&-76.8\%&-89.8\%&-79.1\%&-93.1\%&-35.6\%&-35.2\%
 \\
ResNet-152-P & 0.579 & 0.616 & 0.523 & 0.499 & 0.510 & 0.360 & 0.391 & 0.353 & 0.093 & 0.044 & 0.487 & 0.498 \\
\rowcolor{gray!30}
Gain & -25.9\%&-21.2\%&-33.0\%&-36.2\%&-34.7\%&-54.0\%&-49.9\%&-54.9\%&-88.1\%&-94.4\%&-37.6\%&-36.3\%
 \\
Swin-v2-T-P & 0.586 & 0.629 & 0.507 & 0.453 & 0.480 & 0.254 & 0.265 & 0.263 & 0.158 & 0.145 & 0.464 & 0.468 \\
\rowcolor{gray!30}
Gain & -20.5\%&-13.8\%&-31.2\%&-37.9\%&-34.9\%&-65.2\%&-64.0\%&-64.0\%&-78.6\%&-80.1\%&-37.0\%&-35.9\%
 \\
Swin-v2-B-P & 0.458 & 0.455 & 0.483 & 0.422 & 0.458 & 0.432 & 0.218 & 0.145 & 0.282 & 0.209 & 0.415 & 0.402 \\
\rowcolor{gray!30}
Gain &-16.9\%&-12.0\%&-12.3\%&-18.4\%&-16.9\%&-16.4\%&-60.4\%&-72.0\%&-48.8\%&-59.6\%&-24.7\%&-22.2\%

 \\
MaxViT-T-P & 0.728 & 0.736 & 0.564 & 0.507 & 0.387 & 0.155 & 0.017 & -0.018 & 0.056 & 0.019 & 0.397 & 0.403 \\
\rowcolor{gray!30}
Gain & -5.6\%&-4.4\%&-26.8\%&-34.2\%&-49.8\%&-79.9\%&-97.8\%&-102.3\%&-92.7\%&-97.5\%&-48.5\%&-47.7\%
 \\ 
  Mean Gain & -23.3\%&-19.6\%&-33.1\%&-38.7\%&-37.1\%&-57.3\%&-73.9\%&-79.1\%&-74.8\%&-81.5\%&-36.8\%&-35.7\%
\\
\bottomrule
\end{tabular}
\label{tab:q3_jufe_10k}
\end{table*}

\textbf{$\mathbf{\romannumeral4)}$ Debiased validation.} The analysis results from the first three perspectives are derived from existing PIQA databases, which are inevitably subject to biases introduced by the in-lab environment. To analyze the benefit of each PIQA database in practice, we utilize the BPIQA models initialized on the target database to assess the samples from the real world as shown in Fig.~\ref{fig:analysis_framework} (d). The experimental results are presented in Tab.~\ref{tab:q4}, it is evident that the BPIQA models initialized on the OIQ-10K database exhibit the best performance, indicating that they can extract significantly more distortion information from this database than from others. Meanwhile, OIQA and MVAQD deliver sub-optimal results, yet it is apparent that all three databases feature a diverse array of distortion types and levels, coupled with rich and varied image content. Since CVIQ, IQA-OID, and JUFE-10K are tailored to samples featuring compressed distortion, map projection, and heterogeneous distortion, they serve to enhance BPIQA models in particular aspects but inherently exhibit a certain degree of limited generalization ability. OSIQA yields the worst results, primarily due to its small number of samples and sole focus on stitching distortion. Eventually, we rank all PIQA databases as follows (from highest to lowest performance): OIQ-10K, OIQA, MVAQD, JUFE-10K, CVIQ, IQA-ODI and OSIQA.

\begin{table*}[t]
\centering
\small
\setlength{\tabcolsep}{4pt}
\caption{Comparison of the Results of BPIQA Models Trained on Various Distorted Panoramic Image Databases and Tested on AIGCOIQA.}
\begin{tabular}{lcccccccccccccc}
\toprule
\multirow{2}{*}{Method} & \multicolumn{2}{c}{CVIQ (5)} & \multicolumn{2}{c}{OIQA (2)} & \multicolumn{2}{c}{MVAQD (3)} & \multicolumn{2}{c}{IQA-ODI (6)} & \multicolumn{2}{c}{OSIQA (7)}& \multicolumn{2}{c}{OIQ-10K (1)} & \multicolumn{2}{c}{JUFE-10K (4)} \\
 & PLCC & SRCC & PLCC & SRCC & PLCC & SRCC & PLCC & SRCC & PLCC & SRCC & PLCC & SRCC & PLCC & SRCC\\
\midrule
MC360IQA & 0.259 & 0.254 & 0.603 & 0.534 & 0.116 & 0.023 & 0.213 & 0.222 & 0.203 & 0.136 & 0.550 & 0.518 & 0.393 & 0.378 \\
Assessor360 & 0.249 & 0.218 & 0.608 & 0.555 & 0.638 & 0.596 & 0.188 & 0.140 & 0.196 & 0.213 & 0.687 & 0.682 & 0.130 & 0.132 \\
OIQAND & 0.196 & 0.225 & 0.555 & 0.486 & 0.719 & 0.715 & 0.219 & 0.155 & 0.188 & 0.196 & 0.587 & 0.581 & 0.523 & 0.537 \\
\midrule
ResNet-50-P & 0.509 & 0.498 & 0.532 & 0.467 & 0.593 & 0.573 & 0.286 & 0.143 & 0.126 & 0.088 & 0.678 & 0.667 & 0.417 & 0.447 \\
ResNet-152-P & 0.502 & 0.420 & 0.564 & 0.560 & 0.482 & 0.483 & 0.245 & -0.006 & 0.236 & 0.019 & 0.724 & 0.736 & 0.525 & 0.543 \\
Swin-v2-T-P & 0.308 & 0.272 & 0.620 & 0.594 & 0.708 & 0.704 & 0.211 & 0.183 & 0.020 & 0.042 & 0.686 & 0.676 & 0.378 & 0.412 \\
Swin-v2-B-P & 0.381 & 0.193 & 0.641 & 0.639 & 0.700 & 0.683 & 0.142 & -0.045 & 0.211 & 0.133 & 0.672 & 0.668 & 0.359 & 0.396 \\
MaxViT-T-P & 0.486 & 0.448 & 0.516 & 0.483 & 0.755 & 0.777 & 0.232 & 0.195 & 0.150 & 0.115 & 0.672 & 0.680 & 0.454 & 0.446 \\
\rowcolor{gray!30}
Mean &0.361&0.316&0.580&0.540&0.589&0.569&0.217&0.123& 0.166 & 0.118 & \textbf{0.657}&\textbf{0.651}&0.397&0.393\\
\bottomrule
\end{tabular}
\label{tab:q4}
\end{table*}

\begin{table}[t]
\centering
\caption{The Global Ranking Results of Current PIQA Databases from Best to Worst. Q1, Q2, and Q3 Denote the Ranking Results of the Three Perspectives, Respectively. T Denotes the Ranking Results of the Debiased Validation.}
\begin{tabular}{lccccc}
\toprule
\multirow{2}{*}{Database} & \multicolumn{5}{c}{Global Ranking} \\
  & Q1  & Q2  & Q3 & T & Final\\
\midrule
OIQ-10K & 4 & 2 & 1 & 1 & 1 \\
JUFE-10K & 2 & 1 & 4 & 4 & 2 \\
MVAQD & 5 & 4 & 3  & 3 & 3 \\
OIQA  & 6 & 6 & 2 & 2 & 4 \\
OSIQA  & 1 & 3 & 7 & 7 & 5 \\
IQA-ODI & 3 & 5 & 6  & 6 & 6 \\
CVIQ  & 7 & 7 & 5 & 5 & 7 \\
\bottomrule
\end{tabular}
\label{tab:rank}
\end{table}

\subsection{Analysis Summary}
 \textbf{$\mathbf{\romannumeral1)}$ Global ranking results.} We first empirically rank all PIQA databases based on the analysis of three perspectives and the performance on in-the-wild samples respectively, and then induct the final global ranking results for a comprehensive analysis. Specifically, we consider the four aspects equally important in our study. Therefore, the average ranking score is used to determine the final result. As the results are shown in Tab.~\ref{tab:rank}, \textbf{OIQ-10K} enables BPIQA models to learn the best distortion-related knowledge to assess the perceptual quality of most samples whether they are from in-lab databases or out-of-lab databases. Besides, this database causes a little challenge to the current BPIQA models, leading to an unsaturated performance which helps advance the \textit{ture} of the current status of BPIQA. The success of OIQ-10K can be attributed to its consideration of real-world local degradation, which includes the introduction of \textit{homogeneous distortion} and \textit{heterogeneous distortion} simultaneously. Moreover, it encompasses a diverse range of distortion types, varying degrees of distortion, and various image resolutions, among other factors. The second high-quality PIQA database is \textbf{JUFE-10K}, it successfully creates a big gap between BIQA and BPIQA that disables the BIQA models to access the BPIQA. JUFE-10K focuses predominantly on the common local degradation issues encountered in real-world scenarios, encompassing a handful of panoramic images with \textit{heterogeneous distortions}. These distortions pose significant challenges to BPIQA models, necessitating the introduction of specific design within the models to effectively address this unique distortion pattern. However, it seems that no one BPIQA model has successfully processed this distortion pattern so far, causing JUFE-10K to hardly enable BPIQA models to effectively process all distorted scenarios due to the specificity of the distortions. Although \textbf{MVAQD} attains a relative average rank in this computational analysis, it is worthy to be constructed and deserves further consideration in the future. The main reason is that MVAQD only contains a few samples compared with OIQ-10K and JUFE-10K, and reveals certain beneficial information for building the BPIQA model, suggesting that the diversity of distortion situation and image resolution are of great importance in the PIQA database. \textbf{OIQA} fails to create a gap between BIQA and BPIQA, leading to BIQA models substituting BPIQA models to assess the perceptual quality of panoramic images without any effort, which can not expose the \textit{true} of the current status of BPIQA. In addition, BPIQA models present a saturated performance on this database that causes a waste of the human labeling budget. Surprisingly, the BPIQA models trained on OIQA demonstrate favorable generalization ability, this could attributed to the source images captured by professional photographers with diverse compositions and rich colors. \textbf{OSIQA} yield a polarized result, which lead to a less prominent overall quality. On one hand, it prevents BIQA models from recognizing its distortion patterns and significantly hinders BPIQA models, suggesting that it successfully encourages models to design better modules to handle its distortions. On the other hand, it struggles to improve the model's generalization ability, proving that its distortions lack universal features, thus greatly diminishing its practical value. This result is understandable, similar to JUFE-10K, as both databases only contain heterogeneous distortions. Although \textbf{IQA-ODI} boasts a significantly larger sample size and possesses distinctive characteristics that distinguish it from planar images, unfortunately, it fails to pose effective obstacles for BPIQA models thereby enhancing their assessment capabilities through it. The main reason may be the single distortion type can not offer more distortion-related knowledge. Apparently, \textbf{CVIQ} suffers from a severe \textit{easy-database} issue, which hinders the practical application in BPIQA. The reason is multiple. First, the source images in CVIQ have fewer distinctive features causing the overall samples with insufficient richness and depth. Second, CVIQ centers on compressed distortion, however, the panoramic images in the real world are commonly subjected to various distortions. Besides, the image resolution (only 4K) of samples in CVIQ is relatively sufficient compared to other PIQA databases.
 
 \textbf{$\mathbf{\romannumeral2)}$ The proposed general BPIQA models.} We analyze the experimental results in Tab.~\ref{tab:q2} to explore our proposed general models. Specifically, the average PLCC value of the proposed five models reaches 0.978, 0.963, and 0.964 on CVIQ, OIQA, and IQA-ODI, respectively, while achieving a promising average performance with a PLCC of 0.801 on OIQ-10K. However, compared to OIQAND, the results on the databases with pure heterogeneous distortion (\textit{i.e.,} OSIQA and JUFE-10K) are relatively worse, which indicate that special designs are essential to address these challenging scenarios.

\section{Conclusion and Future Work}
\label{sec:conc}

In this paper, we conduct a thorough analysis of the \textit{easy-database} issue emerging BPIQA from three perspectives, including the gap between BPIQA and BIQA, the necessity of specific design in BPIQA models, and the generalization ability of BPIQA models. Experimental results demonstrate that most of PIQA databases suffer from the \textit{easy-database} issue, and are not sensitive to the specific designs in BPIQA models. Moreover, we study the PIQA databases by the generalization ability of BPIQA models, and we find that a high-quality PIQA database can empower BPIQA to deal with hard samples and exhibit high transfer performance in cross-validation experiments. Finally, we empirically rank all PIQA databases based on the above analysis and induct a global ranking about the \textit{easy-database} issue, which could give insightful guidelines for next-generation PIQA databases and BPIQA models. 

From the database perspective, we can observe that OSIQA, OIQ-10K and JUFE-10K present significant challenges for both BIQA and BPIQA, which implies that there still exists substantial improvement room for PIQA in real-world applications. Inspired by the results, we can put more subjective effort on those \textit{hard samples} (\emph{i.e.}, images with non-uniform distortion distribution), and then design more powerful objective models by incorporating the characteristic of how human beings perceive these images. In addition, we can also shift partial attention to those panoramic images degraded by processing algorithms~\cite{yan2023survey}. From the model perspective, the transformer-based and hybrid-based models are worth further deep exploration, since the ability of global modeling in the transformer is of great importance in addressing panoramic-specific distortion. We can also consider introducing multimodal information to assist BPIQA models~\cite{zhang2023blind}.




\bibliographystyle{ACM-Reference-Format}
\bibliography{tomm}


\end{document}